\pdfoutput=1
\PassOptionsToPackage{hyphens}{url}
\documentclass[11pt,onecolumn]{cleantechnicalreport}

\usepackage[authoryear,sort&compress,round]{natbib}
\usepackage{hybridfrontmatter}
\usepackage{amsmath,amsfonts,bm}

\def\eqref#1{equation~\ref{#1}}

\def\1{\bm{1}}

\DeclareMathAlphabet{\mathsfit}{\encodingdefault}{\sfdefault}{m}{sl}
\SetMathAlphabet{\mathsfit}{bold}{\encodingdefault}{\sfdefault}{bx}{n}

\usepackage{multirow}
\usepackage{wrapfig}
\usepackage{placeins}
\usepackage{fontawesome5}

\DeclareUnicodeCharacter{2011}{-}
\DeclareUnicodeCharacter{2019}{'}

\definecolor{insightbg}{HTML}{EFF6FC}
\definecolor{insighttitlebg}{HTML}{2B5F8A}

\newtcolorbox{insightbox}[1][]{
  enhanced,
  colback=insightbg,
  colframe=insighttitlebg,
  left=4mm, right=4mm, top=3mm, bottom=3mm,
  fonttitle=\bfseries\color{white},
  title={\scalebox{0.7}{\faLightbulb}~~#1},
  attach boxed title to top left={xshift=3mm, yshift=-3mm},
  boxed title style={colback=insighttitlebg, sharp corners, size=small},
  arc=3mm, boxrule=0.8pt,
}

\renewcommand{\ReportTitleFont}{%
  \normalfont\bfseries\fontsize{14}{17}\selectfont}

\makeatletter
\fancypagestyle{hybridfirststyle}{%
  \fancyhf{}%
  \fancyfoot[L]{%
    \ifdefempty{\report@footertext}{}{%
      {\color{black}\normalfont\fontsize{8}{10}\selectfont\report@footertext}%
    }%
  }%
  \fancyfoot[C]{}%
  \renewcommand{\footrule}{\color{gray}\DefaultFootRule}%
}
\makeatother

\hypersetup{
  colorlinks=true,
  linkcolor=blue,
  citecolor=blue,
  urlcolor=blue,
  pdftitle={An Empirical Study of Parameter-Efficient Fine-Tuning for Mixture-of-Experts LLMs},
  pdfauthor={Minglei Li, Peng Ye, Haonan He, Jingqi Ye, Qingyu Yang, Tao Chen}
}

\let\cite\citep
\title{An Empirical Study of Parameter-Efficient Fine-Tuning for Mixture-of-Experts LLMs}
\author{%
\begin{minipage}{0.98\textwidth}
\centering
\normalfont
{\small
\textbf{Minglei Li}\textsuperscript{1}\ReportAuthorMark{*}\quad
\textbf{Peng Ye}\textsuperscript{2,3}\ReportAuthorMark{*}\quad
\textbf{Haonan He}\textsuperscript{2,4}\quad
\textbf{Jingqi Ye}\textsuperscript{2,4}\quad
\textbf{Qingyu Yang}\textsuperscript{2,5}\quad
\textbf{Tao Chen}\textsuperscript{1,6}\ReportAuthorMark{\ensuremath{\dagger}}}
\end{minipage}%
}
\affil{%
\textsuperscript{1}\textbf{Fudan University}\quad
\textsuperscript{2}\textbf{Shanghai Artificial Intelligence Laboratory}\\
\textsuperscript{3}\textbf{The Chinese University of Hong Kong}\quad
\textsuperscript{4}\textbf{University of Science and Technology of China}\\
\textsuperscript{5}\textbf{KTH Royal Institute of Technology}\quad
\textsuperscript{6}\textbf{Shanghai Innovation Institute}
}
\reportauthornote{\ReportAuthorMark{*} Equal contribution.\quad
\ReportAuthorMark{\ensuremath{\dagger}} Corresponding author.}
\date{}
\reportwebsite{https://leoli646.github.io/moefactory-project-page/}
\reportcode{https://github.com/leoli646/MoEFactory}
\reportdate{October, 2026}

\begin{abstract}
Parameter-efficient fine-tuning (PEFT) methods are critical for adapting large models under resource-constrained settings. However, as mixture-of-experts (MoE) architectures, which offer efficient inference at massive parameter scales, gain increasing prevalence in the community, a critical gap arising from the lack of systematic recipes for applying PEFT to MoE models. To narrow this gap, we conduct a large-scale systematic study that investigates a set of essential design recipes, including which modules should be adapted, which granularity should be used to adapt MoE modules, which PEFT method to use, and to what extent should the load-balancing objective be used, across three MoE backbones on downstream tasks and general knowledge tasks. Our experiments yield several key insights.  First, adapting MoE experts alone achieves the highest downstream performance with competitive general capability retention under a comparable parameter budget. Second, native expert-weight updates offer a better downstream--retention trade-off than adapters that bypass experts or entire MoE modules. Third, LoRA and DoRA together define the strongest operating region for high downstream accuracy among the evaluated PEFT methods. Fourth, no nonzero load-balancing coefficient Pareto-dominates the zero baseline.
We finally evaluate a weight-level redesign of MoE$^2$-LoRA on Qwen3. Moving its module-level bypass pool into native expert weights
improves all four upstream endpoints by 2.4 to 17.9 points with 17\% fewer trainable parameters and a 3.2-point average downstream cost.
\end{abstract}

\begin{document}
\vspace*{-\headheight}
\maketitle

\section{Introduction}

Mixture-of-experts architectures~\citep{shazeer2017sparsely,lepikhin2021gshard,fedus2022switch,du2022glam} enable extensive scaling of model parameters while activating only a small fraction of parameters per input unit. This sparse activation paradigm offers an outstanding trade-off between performance and computation, and has consequently been adopted by frontier models such as GLM‑5.2~\citep{glm5team2026glm5vibecodingagentic} and Kimi‑K3~\citep{kimiteam2026kimik3openfrontier}. Despite the increasing prevalence of MoE architectures, most existing parameter‑efficient fine‑tuning (PEFT) methods~\citep{hu2022lora,houlsby2019adapters,li2021prefix} and recipes were originally designed for dense models, and the community still lacks a systematic guideline for applying PEFT methods to MoE structures. Given that MoE models typically comprise a massive number of parameters, it is essential to adopt a scientifically grounded guideline for applying PEFT to such models to achieve efficient adaptation. This gap therefore raises a pressing question: \emph{How can we adapt MoE models to achieve the best performance-efficiency balance with the least forgetting?}

To answer this question, we conduct a large-scale systematic empirical study using representative MoE backbones from different families, including Qwen3-30B-A3B-Base~\citep{qwen2025qwen3}, Gemma-4-26B-A4B~\citep{gemmateam2026gemma4technicalreport}, and GPT-OSS-20B~\citep{openai2025gptoss20b}. Specifically, we structure our study around four main axes: (1)~which modules should be targeted for adapting MoE models under a similar trainable parameter budget? (2)~should adaptation granularity for MoE modules be set at the module level, expert level, or weight level? (3)~which PEFT method generally works best for adapting MoE models? (4)~how strong should the auxiliary loss coefficient be set for adapting MoE models? We conduct the study using two distinct downstream tasks and two general tasks. The first downstream task is mathematical reasoning, on which base models typically already possess strong capabilities; it serves to probe whether further leverages well-established knowledge. The second downstream task is an intent classification task, which yields unsatisfactory results for the three base models and therefore provides a more stringent test of a PEFT method’s ability to adapt in under-represented regimes.

Our study therefore yields four instructive insights: (1)~applying adaptation solely to experts, achieves the best downstream performance; (2)~within the MoE modules, applying adaptation at the weight level is more reliable than at the expert level or module level; (3)~among the evaluated PEFT methods, LoRA and DoRA deliver the strongest downstream performance, whereas methods with weaker general capabilities retention can fail to learn the target task properly; (4)~auxiliary load balancing may degrade performance on both downstream tasks and general capabilities retention. Guided by these findings, we revisit MoE$^2$-LoRA, a PEFT method tailred for MoE fine-tuning, which established a learned mapping between the selection of pre-trained MoE experts and that of LoRA experts.
Our modification on MoE$^2$-LoRA improves all four upstream endpoints by 2.4 to 17.9 points with 17\% fewer trainable parameters and only a 3.2-point average downstream cost, turning a large task-dependent forgetting into near-complete retention and thereby showing that our insights are effective beyond the controlled grid.

Our contributions are summarized as follows:
\begin{itemize}[left=0pt,labelsep=1em]
\item To our knowledge, this is the first large-scale PEFT study conducted directly on pretrained MoE LLMs, covering four design axes, three heterogeneous backbones, and two downstream tasks. Beyond the downstream accuracy that PEFT work usually reports alone, we pair every selected checkpoint with general-capability evaluation on MMLU-Pro and IFEval, so each design choice is judged by what it gains and by what it costs the pretrained model.
\item We distill the study into four actionable insights on adaptation position, expert granularity, update form, and load balancing, each stated with the regime in which it holds, providing concrete defaults and open questions for future work on MoE adaptation. Functional and spectral analyses explain how weight-level updates reuse pretrained expert computation.
\item We show the insights are practical by refining MoE$^2$-LoRA, a recent routed-pool method, along the single axis it does not already follow, which improves all four general-capability endpoints with 17\% fewer parameters.
\end{itemize}

\section{Related Work}

\paragraph{Parameter-efficient Fine-tuning (PEFT).}

PEFT adapts large models by learning a small set of parameters while keeping pretrained weights largely frozen. Low-rank methods modify pretrained weights through compact updates: LoRA learns a low-rank residual, while DoRA decouples weight magnitude and direction to improve the update~\citep{hu2022lora,liu2024dora}. OFT and FourierFT further restrict the update space through orthogonal transformations and frequency-domain parameterization, respectively~\citep{qiu2024oft,gao2024fourierft}, whereas IA3 adapts the model by scaling intermediate activations~\citep{liu2022tfew}. A complementary line introduces conditional computation into PEFT by organizing prompts, adapters, or LoRA modules as routed experts~\citep{dou2023loramoe,li2024mixlora,gao2024mola,tian2024hydralora,luo2024moelora}. However, these methods are predominantly validated on dense models. It therefore remains unclear how PEFT should interact with the existing experts and routers of an MoE model. In this paper, we systematically study adaptation position, expert granularity, update form, and load balancing to derive principled guidelines for applying PEFT to MoE models.

\begingroup
\setlength{\intextsep}{6pt}
\begin{wrapfigure}{R}{0.45\textwidth}
\centering
\includegraphics[width=\linewidth]{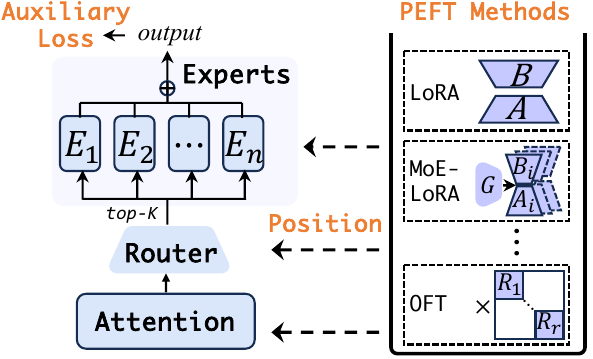}
\captionsetup{font=small,skip=4pt}
\caption{Illustration of Position, PEFT Methods, and Auxiliary Loss.}
\label{fig:pipeline-design}
\end{wrapfigure}

\paragraph{PEFT for MoE architecture models.}
PEFT for pretrained MoE models is not a straightforward extension of dense-model PEFT, because sparse routing governs how limited trainable parameters receive supervision. Since each token is dispatched to a few experts, PEFT modules receive uneven, task-dependent signals, making expert selection and update sharing central design choices~\citep{wang2024esft}. Conventional top-$k$ routing also provides incomplete gradients for PEFT modules optimization during post-training~\citep{yao2025densemixer}. Moreover, load-balancing objectives may enforce overly uniform routing and weaken the specialization induced by expert-specific updates on narrow downstream distributions~\citep{wang2024auxiliary,guo2025specialization}. Existing PEFT studies address parts of this through task-relevant expert selection, expert-wise allocation, joint router--expert adaptation, or additional routed adaptation modules~\citep{wang2024esft,bai2025understanding,deng2026dr,liu2024perft,yang2026moe2lora}. However, they often vary simultaneously in adaptation position, expert granularity, update form, routing strategy, and balancing objective, making it difficult to determine which choices consistently benefit. We therefore conduct a controlled study of these factors to derive robust principles for MoE adaptation.

\section{Background and Study Design}

\subsection{Background}

\begin{wrapfigure}{r}{0.45\textwidth}
\vspace{-\intextsep}
\centering
\includegraphics[width=\linewidth]{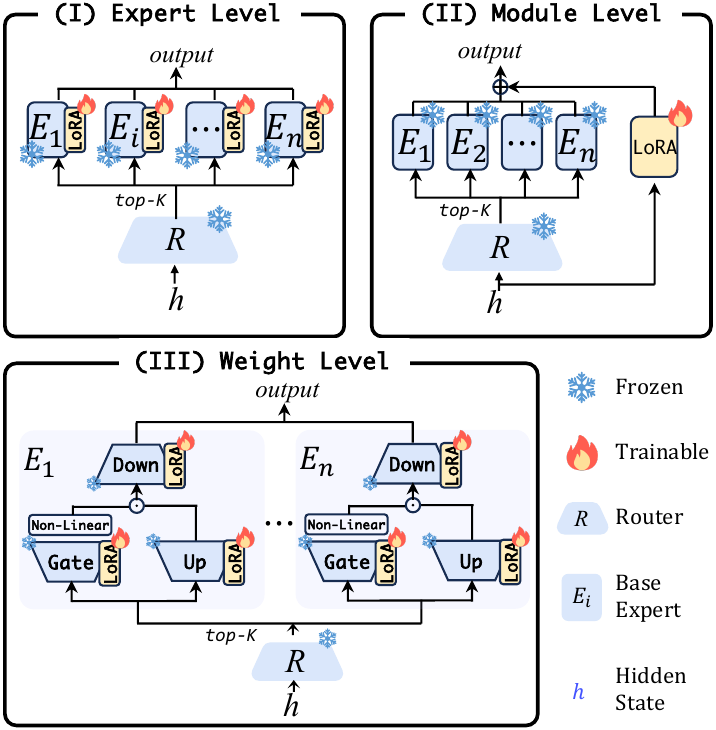}
\captionsetup{font=small,skip=4pt}
\caption{Three evaluated adaptation granularities of the MoE module.}
\label{fig:granularity-design}
\end{wrapfigure}

This work investigates the principled application of PEFT methods to MoE models, with an emphasis on the MoE modules.
Formally, given a hidden state \(x\), the computation inside an MoE module can be written as
\begin{equation}
    y = \sum_{e \in \operatorname{TopK}(g(x))} p_e(x) \, f_e(x),
\end{equation}
where \(g\) denotes the router, \(p_e\) is the routing probability for expert \(e\), and \(f_e\) represents the transformation of expert \(e\).

We model the fine-tuning process as an incremental update of the pre-trained weights:
\begin{equation}
    W' = W + \Delta W,
\end{equation}
where \(W' \in \mathbb{R}^{d_\mathrm{out} \times d_\mathrm{in}}\) is the fine-tuned weight, and \(W\) and \(\Delta W\) correspond to the pre-trained weight and its additive update, respectively.
In the case of LoRA, the update is parameterized as \(\Delta W = \frac{\alpha}{r} B A\), with low-rank matrices \(A \in \mathbb{R}^{r \times d_\mathrm{in}}\) and \(B \in \mathbb{R}^{d_\mathrm{out} \times r}\), while \(\alpha\) and \(r\) are hyper-parameters controlling the update scale and rank.
Other PEFT methods evaluated in this study are deferred to Appendix~\ref{app:overview}.

\par\WFclear
\endgroup

\subsection{Study Design}


We evaluate Qwen3-30B-A3B-Base, Gemma-4-26B-A4B, and GPT-OSS-20B on two complementary downstream tasks, specifically mathematical reasoning and intent classification. Appendix~\ref{app:backbones} details their architectures. For the intent classification task, the models are trained on ESFT\_Intent~\citep{wang2024esft} and evaluated by exact match on 500 held-out examples. For the mathematical reasoning task, models are trained on MetaMathQA-R1, derived from MetaMathQA~\citep{yu2024metamath}, and evaluated on MATH-500~\citep{hendrycks2021math}.

We use MMLU-Pro~\citep{wang2024mmlupro} and IFEval~\citep{zhou2023ifeval} to measure the general capabilities, including knowledge-based reasoning and instruction-following retained after downstream adaptation. For each configuration and model--task pair in the four design studies, we train three seeds per candidate learning rate and select the rate with the highest mean downstream score.
We evaluate the three checkpoints at that rate on MMLU-Pro and IFEval and report three-seed means, without using general scores for selection. \textbf{Detailed implementation settings are provided in Appendix~\ref{app:protocol}.} Selected learning rates and Qwen3 training throughput are reported in Appendix~\ref{app:qwen3-training-throughput}.

\section{Empirical Study}
\label{sec:empirical}

\subsection{Which Modules Should be Targeted for Adaptation?}

We compare attention-only ($r=64$), expert-only ($r=256$), and attention+experts ($r=32$ and $128$) under a comparable budget, using one shared branch for the expert block to keep the focus on position. We repeat each with a rank-8 router update and also test the higher-budget attention/expert pair ($r=64/256$), with and without router adaptation.

\begin{figure}[!t]
\centering
\includegraphics[width=1.0\linewidth]{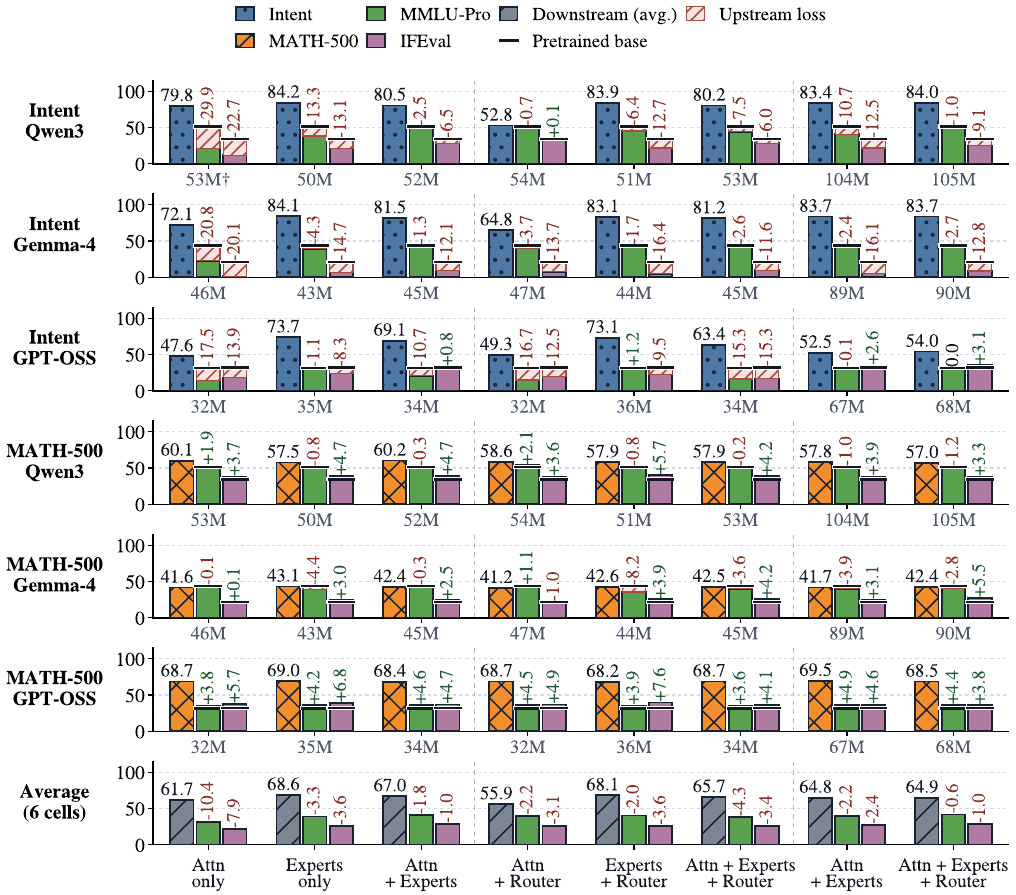}
\caption{Visualized results of the study of adapted modules. Attn denotes attention, and gray labels report trainable parameters.}
\label{fig:position}
\end{figure}

Figure~\ref{fig:position} summarizes the adaptation-position study. Expert-only adaptation leads across the six model--task cells at 68.6\%, ahead of experts+router at 68.1\% and attention+experts at 67.0\%. It changes MMLU-Pro and IFEval by $-3.3$ and $-3.6$ points. Attention-only reaches 61.7\%, with changes of $-10.4$ and $-7.9$ points. Attention+experts changes the two benchmarks by $-1.8$ and $-1.0$ points, while experts+router changes them by $-2.0$ and $-3.6$ points. Without and with router adaptation, the higher-budget settings reach 64.8\% and 64.9\%; MMLU-Pro changes by $-2.2$ and $-0.6$ points, and IFEval by $-2.4$ and $-1.0$ points. Expert-only is strongest at comparable budget, while attention+experts retains more general capability.

\begin{insightbox}[Insight 1.]
For MoE fine-tuning, target the experts by default, and adapt the router only when the task calls for routing specialization.
\end{insightbox}

\subsection{Which Granularity Should be Used to Adapt MoE Modules?}

Figure~\ref{fig:granularity-design} contrasts a rank-256 module-level bypass, a rank-2 expert-level bypass, and weight-level updates at ranks 1 and 2.

\begin{figure}[!t]
\centering
\includegraphics[width=1.0\linewidth]{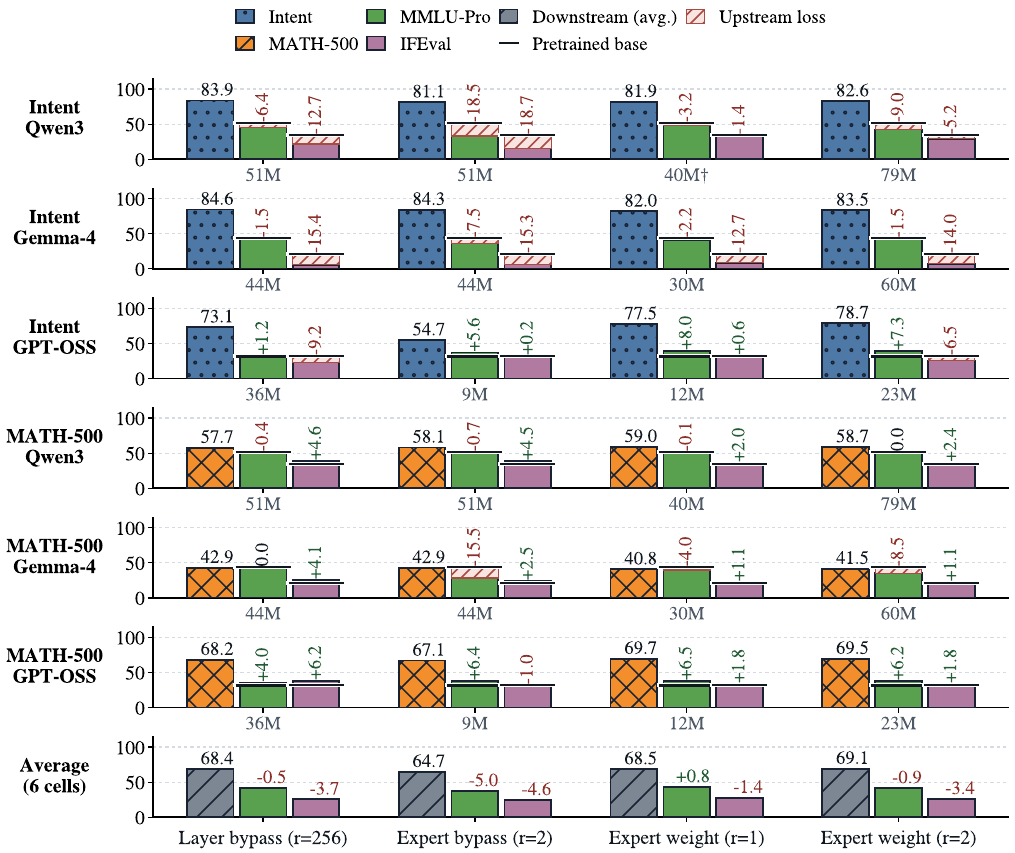}
\caption{Visualized results of the study of MoE adaptation granularity.}
\label{fig:granularity}
\end{figure}

Figure~\ref{fig:granularity} shows weight-level adaptation leading at rank 2 with 69.1\%, followed by rank 1 at 68.5\%, module-level bypass at 68.4\%, and expert-level bypass at 64.7\%. Rank 1 has the strongest general-capability profile, changing MMLU-Pro and IFEval by $+0.8$ and $-1.4$ points; rank 2 changes them by $-0.9$ and $-3.4$. Both bypass alternatives reduce both benchmarks. Rank 2 gains 0.6 downstream points at roughly twice the trainable-parameter count.

\begin{insightbox}[Insight 2.]
Weight-level adaptation provides the best trade-off. Rank-$1$ is a strong default, achieving near-optimal downstream performance while preserving general capabilities. Increasing the rank brings only marginal gains, whereas bypass-based adaptation is generally less stable and less effective.
\end{insightbox}

\subsection{Which PEFT Method Generally Works Best for Adapting MoE Models?}

We compare LoRA~\citep{hu2022lora}, DoRA~\citep{liu2024dora}, OFT~\citep{qiu2024oft}, FourierFT~\citep{gao2024fourierft}, and IA3~\citep{liu2022tfew}, alongside expert-level LoRA, MoE-LoRA~\citep{luo2024moelora}, module-level LoRA, DAS-LoRA~\citep{tang2026exploring}, and last-layer partial fine-tuning~\citep{ye2023partialfinetuningsuccessorfinetuning}.

\begin{figure}[!t]
\centering
\includegraphics[width=1.0\linewidth]{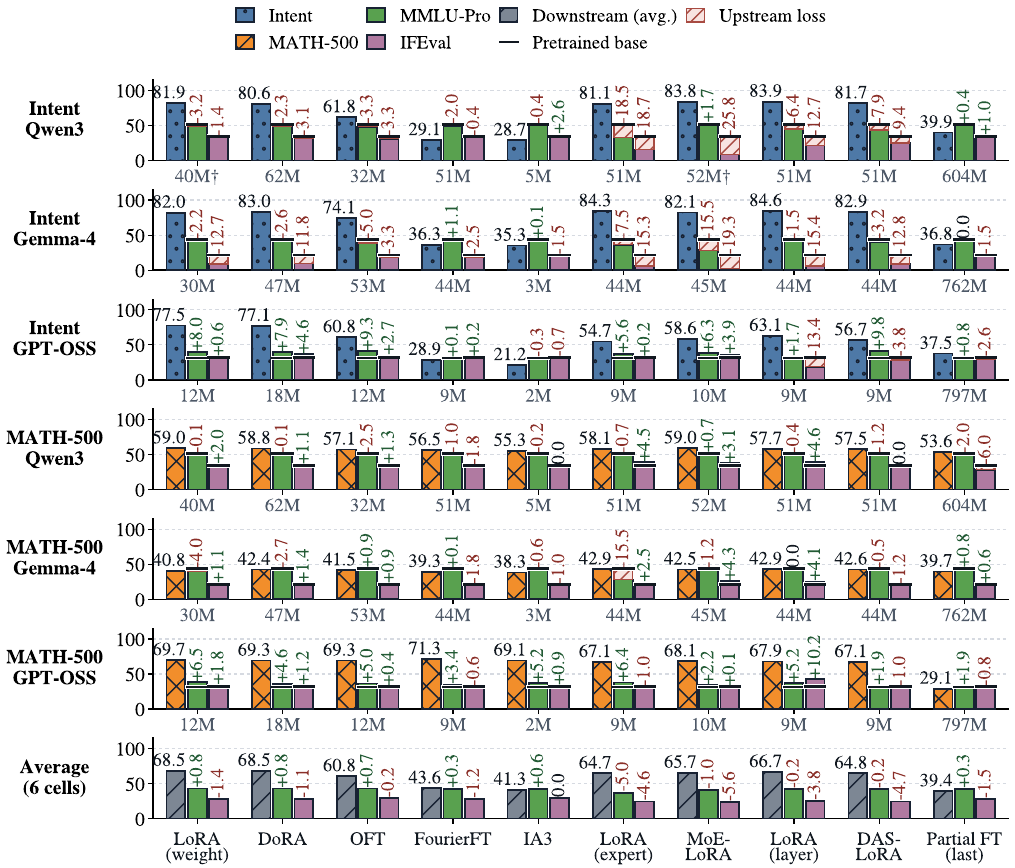}
\caption{Visualized results of the study of PEFT methods.}
\label{fig:update}
\end{figure}

Weight-level LoRA and DoRA tie at 68.5\% across the six downstream cells. Both raise MMLU-Pro by 0.8 points; DoRA changes IFEval by -1.1, compared with -1.4 for LoRA. Among MoE-specific variants, layer-level LoRA leads at 66.7\%, followed by MoE-LoRA at 65.7\%, DAS-LoRA at 64.8\%, and expert-level LoRA at 64.7\%. OFT, FourierFT, IA3, and partial fine-tuning reach 60.8\%, 43.6\%, 41.3\%, and 39.4\%. Weight-level LoRA and DoRA provide the strongest downstream accuracy, with DoRA retaining more IFEval performance.

\begin{insightbox}[Insight 3.]
Start expert-weight adaptation with standard LoRA or DoRA. Favor DoRA when preserving instruction-following performance is a priority.
\end{insightbox}

\subsection{How Strong Should the Auxiliary Loss Coefficient Be Set For Adapting MoE Models?}

\begin{figure}[!t]
\centering
\includegraphics[width=1.0\linewidth]{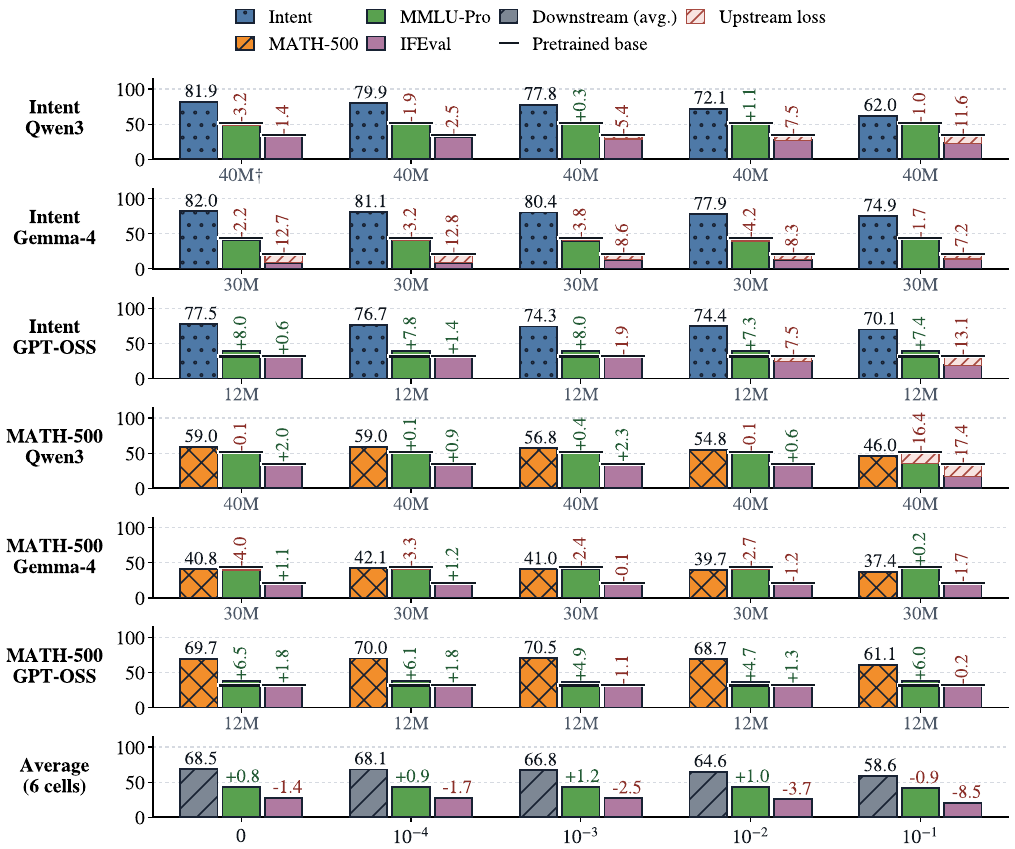}
\caption{Visualized results of the study of the strength of auxiliary loss.}
\label{fig:aux}
\end{figure}

We vary the auxiliary loss coefficient from zero to $10^{-1}$ while holding the adaptation method and trainable-parameter count fixed, as illustrated in Figure~\ref{fig:aux}.

Figure~\ref{fig:aux} shows a 68.5\% downstream average without auxiliary loss; coefficients of $10^{-4}$, $10^{-3}$, $10^{-2}$, and $10^{-1}$ yield 68.1\%, 66.8\%, 64.6\%, and 58.6\%. At $10^{-3}$, MMLU-Pro rises 1.2 points, IFEval falls 2.5, and the downstream average falls 1.7 relative to zero. At $10^{-1}$, MMLU-Pro and IFEval fall 0.9 and 8.5 points. Larger coefficients therefore reduce downstream performance and IFEval retention.

\begin{insightbox}[Insight 4.]
The auxiliary loss should be disabled or assigned a very small coefficient during MoE adaptation. A coefficient around $10^{-4}$ is a safe choice when routing regularization is required, whereas larger values can substantially degrade downstream performance.
\end{insightbox}

\section{Design Implications}
\label{sec:implications}

To test whether the insights transfer to a composite method, we refine MoE$^2$-LoRA~\citep{yang2026moe2lora}. It follows Insight~1 by adapting only MoE modules with frozen attention and follows Insight~4 by disabling auxiliary load balancing. It also uses the native router to select entries from a shared pool without updating router weights. Its remaining mismatch is Insight~2: the pool is a module-level bypass rather than an update to native expert projections. We therefore replace each of the 128 rank-32 pool entries with a low-rank update to native expert weights while retaining the routed pool, router-driven selection, frozen attention, and zero balancing coefficient.
Appendix~\ref{app:moe2} gives both update equations.

\begin{wraptable}{r}{0.52\textwidth}
\vspace{-\intextsep}
\small
\centering
{
\setlength{\tabcolsep}{3pt}
\begin{tabular}{@{}llrr@{}}
\toprule
Checkpoint & Benchmark & Original & Weight Level \\
\midrule
\multicolumn{2}{@{}l}{Trainable params} & 47.71M & \textbf{39.58M} \\
\midrule
\multirow{3}{*}{Math} & MATH-500 & \textbf{60.0} & 55.8 \\
 & MMLU-Pro & 48.19 & \textbf{50.63} \\
 & IFEval & 35.12 & \textbf{40.30} \\
\midrule
\multirow{3}{*}{Intent} & Intent & \textbf{84.0} & 81.8 \\
 & MMLU-Pro & 45.26 & \textbf{51.20} \\
 & IFEval & 18.11 & \textbf{36.04} \\
\bottomrule
\end{tabular}
}
\setlength{\abovecaptionskip}{4pt}
\caption{Original MoE$^2$-LoRA versus the insight-guided expert-weight redesign on Qwen3.
}
\label{tab:moe2comparison}
\end{wraptable}

Table~\ref{tab:moe2comparison} shows this change reduces the parameter count by 17.0\%, from 47.71M to 39.58M. It improves MMLU-Pro and IFEval by 2.44 and 5.18 points on Math, and by 5.94 and 17.93 points on Intent. The mean absolute difference from the base falls from 6.64 to 2.19 points, and the worst loss from 16.27 to 0.87. Intent remains near base on MMLU-Pro at $-0.30$ and exceeds base on IFEval by $1.66$ points; empty responses stay low and truncation falls from 49.2\% to 11.1\%. The redesign costs 3.2 downstream points while preserving high downstream performance.
Appendix~\ref{app:moe2-validation} reports additional MoE$^2$-LoRA experiments showing that adding attention adaptation or auxiliary load balancing does not consistently improve performance across tasks.

\section{Analysis of Adaptation Granularity}
\label{sec:analysis}
We explain the advantage of weight-level adaptation in Insight 2 through functional response diversity and consistency with pretrained expert structure.
We use RMSNorm-scaled probes to analyze functional response diversity and pretrained spectral consistency in Qwen3 Math and Intent adapters.

\subsection{Functional Response Diversity}
\label{sec:analysis-functional}

For a layer with base output $F_0$ and adapted output $F_a$, collect the changes on shared probes as $Y=[F_a(x_j)-F_0(x_j)]_{j=1}^{n}$. We measure how broadly response energy spans output directions using the functional effective rank
\begin{equation}
  r_{\mathrm{eff}}(Y)
  =\frac{(\operatorname{tr}K)^2}{\lVert K\rVert_F^2+\varepsilon},
  \qquad K=\frac{1}{n}YY^{\top}.
  \label{eq:analysis-effective-rank}
\end{equation}
Equal energy in $k$ orthogonal directions gives $r_{\mathrm{eff}}=k$ up to the numerical stabilizer, making the four granularities comparable.

Weight-level updates reuse the frozen expert's nonlinear features. For a gated expert $f_e(x)=D_e[\sigma(G_ex)\odot U_ex]$, a rank-one up-projection update $\Delta U_e=ab^{\top}$ gives the exact correction
\begin{equation}
  \delta_U f_e(x)=(b^{\top}x)D_e\operatorname{Diag}(a)\sigma(G_ex).
  \label{eq:analysis-rank-one-response}
\end{equation}
The frozen nonlinear features $\sigma(G_ex)$ let a rank-one update span multiple output directions, while routing combines expert-specific responses. Appendix~\ref{app:functional-rank} gives the derivation, rank bounds, and probe protocol.

\begin{figure}[!ht]
\centering
\includegraphics[width=1.0\linewidth]{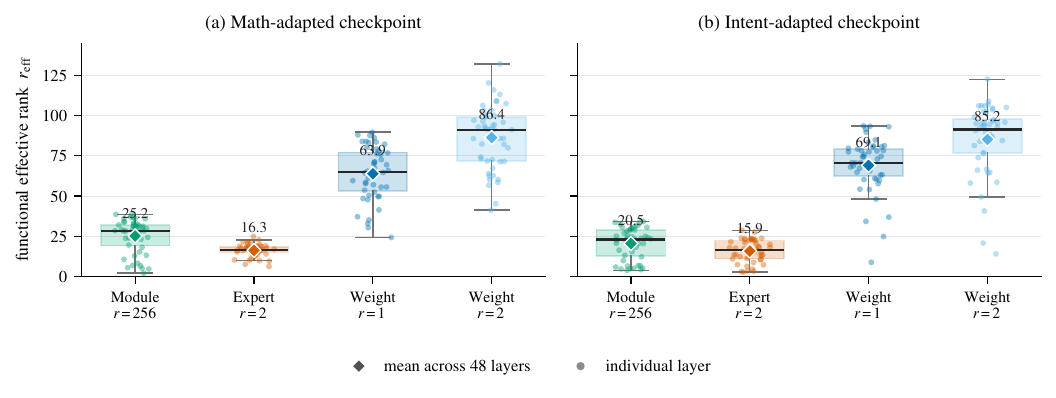}
\caption{Functional response diversity across all 48 Qwen3 MoE layers on synthetic probes. Points denote layers, boxes show medians and interquartile ranges, and diamonds mark means. Weight-level rank-one updates produce broad layer responses on both adapted checkpoints.}
\label{fig:functional-effective-rank}
\end{figure}

Figure~\ref{fig:functional-effective-rank} shows mean effective ranks of 63.90 and 69.08 for weight-level $r=1$ on Math and Intent, compared with 16.31 and 15.91 for expert-level $r=2$. The module-level $r=256$ update reaches 25.25 and 20.54, while weight-level $r=2$ reaches 86.37 and 85.22. Native weight updates therefore produce broad responses at low rank, with additional rank expanding coverage.

\subsection{Pretrained Spectral Consistency}
\label{sec:analysis-spectral}

To test alignment with the pretrained expert, let $u_{e,i}$ and $\sigma_{e,i}$ denote the left singular vectors and descending singular values of its frozen down projection. For perturbations $Y_e$, define the cumulative energy $S_k$ and pretrained profile $S^0_{k,e}$ as
\begin{equation}
  S_k=\frac{\sum_{i\leq k}\lVert u_{e,i}^{\top}Y_e\rVert_2^2}
                {\sum_{i\leq m}\lVert u_{e,i}^{\top}Y_e\rVert_2^2+\varepsilon},
  \qquad
  S^0_{k,e}=\frac{\sum_{i\leq k}\sigma_{e,i}^2}{\sum_{i\leq m}\sigma_{e,i}^2}.
  \label{eq:analysis-spectral-profile}
\end{equation}
At cutoffs $\mathcal{K}=\{16,64,256\}$ and expert width $m=768$, the spectral-consistency score is
\begin{equation}
  C_{\mathrm{spec}}=1-
  \frac{\sum_{k\in\mathcal{K}}|S_k-S^0_{k,e}|}
       {\sum_{k\in\mathcal{K}}|k/m-S^0_{k,e}|+\varepsilon}.
  \label{eq:analysis-spectral-score}
\end{equation}
A score of one matches the pretrained profile, while zero denotes uniform energy. Appendix~\ref{app:spectral-consistency} gives the derivation and full spectral and output-subspace results.

\begin{center}
\includegraphics[width=0.90\linewidth]{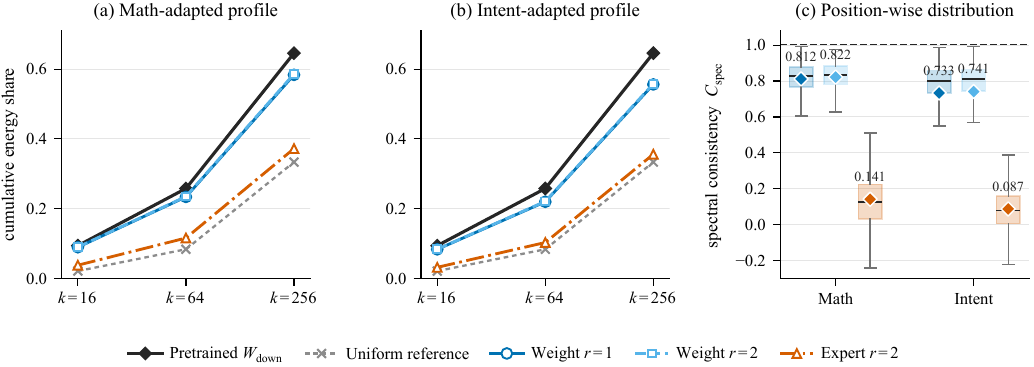}
\captionsetup{font=small,skip=4pt}
\captionof{figure}{Pretrained spectral consistency on Qwen3: Math and Intent energy profiles and score distributions across layer--expert positions. Weight-level updates follow the pretrained spectrum.}
\label{fig:spectral-consistency}
\end{center}

Figure~\ref{fig:spectral-consistency} shows mean $C_{\mathrm{spec}}$ values of $0.812$ and $0.822$ on Math and $0.733$ and $0.741$ on Intent for weight-level ranks 1 and 2, versus $0.141$ and $0.087$ for expert-level rank 2. Rank-one weight updates lead across all 48 layer means and at 99.8\% of Math and 96.3\% of Intent positions. Measurements across gate, up, and down projections confirm spectral alignment; rank two broadens responses while retaining it. These results explain the low-rank efficiency behind Insight~2.

\section{Conclusion}

MoE-PEFT is not only a choice of update parameterization; placement, granularity, optimization, and routing objectives materially change its reliability. Across three backbones and two tasks, expert FFNs are the downstream-robust placement, native expert weights are the most reliable evaluated granularity, LoRA/DoRA lead when a high downstream floor is enforced, and zero auxiliary load balancing is the only no-tuning joint default. None is universal: mixed placement can drift less, low-drift update forms can adapt poorly, and a moderate auxiliary coefficient can trade downstream accuracy for better retention of some general capabilities. Our Qwen3 analyses show that weight-level adaptation combines broad functional responses with strong reuse of pretrained spectral structure.
These defaults also carry over beyond our own grid: a weight-level redesign of MoE$^2$-LoRA that moves its module-level bypass pool into native expert weights improves all four upstream endpoints by 2.4 to 17.9 points with 17\% fewer trainable parameters at a small downstream cost. We therefore offer these defaults as a scoped but directionally reliable starting point for MoE adaptation.
We hope our findings and analyses can stimulate further research on parameter-efficient adaptation of MoE models, and encourage future methods to report what each design choice costs the pretrained model alongside what it gains.

\bibliography{references}
\clearpage
\appendix

\begin{center}
{\large\bfseries An Empirical Study of Parameter-Efficient Fine-Tuning for Mixture-of-Experts LLMs\par}
\vspace{5pt}
{\large Supplementary Material\par}
\end{center}

\vspace{10pt}
The supplementary material is organized as follows:
\begin{itemize}
    \item Appendix~\ref{app:overview} defines the MoE and PEFT updates, including MoE$^2$-LoRA and its redesign.
    \item Appendix~\ref{app:protocol} describes the datasets, checkpoint selection, and evaluation protocol.
    \item Appendix~\ref{app:settings} gives the settings for the four design studies.
    \item Appendix~\ref{app:moe2-validation} reports additional MoE$^2$-LoRA placement and balancing experiments.
    \item Appendix~\ref{app:backbones} compares the backbones' attention and expert architectures.
    \item Appendix~\ref{app:functional-rank} derives and measures functional response diversity across adaptation granularities.
    \item Appendix~\ref{app:spectral-consistency} analyzes pretrained spectral structure and expert output-subspace alignment.
    \item Appendix~\ref{app:qwen3-training-throughput} reports selected learning rates and Qwen3 training throughput.
\end{itemize}

\clearpage

\section{Overview and Notation}
\label{app:overview}

Let $x\in\mathbb{R}^{d}$ denote the token representation entering an MoE
layer.  Each routed expert is a gated feed-forward network.  Using the fused
gate/up projection employed by the evaluated backbones, expert $e$ is
\begin{align}
  \begin{bmatrix}
    a_e(x)\\
    u_e(x)
  \end{bmatrix}
  &= W_{\mathrm{gu},e}x,
  \qquad
  W_{\mathrm{gu},e}
  \in\mathbb{R}^{2d_{\mathrm{ff}}\times d},
  \label{eq:expert-gate-up}\\
  h_e(x)
  &= \phi\!\left(a_e(x)\right)\odot u_e(x),
  \label{eq:expert-hidden}\\
  f_e(x)
  &= W_{\mathrm{down},e}h_e(x),
  \qquad
  W_{\mathrm{down},e}
  \in\mathbb{R}^{d\times d_{\mathrm{ff}}},
  \label{eq:expert-mlp}
\end{align}
where $\phi$ is the backbone activation function; for a SwiGLU expert,
$\phi$ is SiLU.  Equivalently,
$W_{\mathrm{gu},e}$ may be written as separate gate and up matrices
$W_{\mathrm{gate},e}$ and $W_{\mathrm{up},e}$.  Biases are omitted for
clarity.

Let $g(x)\in\mathbb{R}^{E}$ denote the native router logits over $E$ routed
experts.  For a router that renormalizes its selected probabilities, define
\begin{align}
  \bar p_e(x)
  &= \frac{\exp(g_e(x))}
          {\sum_{j=1}^{E}\exp(g_j(x))},
  \\
  \mathcal{S}(x)
  &= \operatorname{TopK}\!\left(\bar p(x),k\right),
  \\
  p_e(x)
  &= \frac{\bar p_e(x)}
          {\sum_{j\in\mathcal{S}(x)}\bar p_j(x)},
  \qquad e\in\mathcal{S}(x).
  \label{eq:topk-router}
\end{align}
The routed MoE output is
\begin{equation}
  y_{\mathrm{routed}}(x)
  = \sum_{e\in\mathcal{S}(x)} p_e(x)f_e(x).
  \label{eq:moe-output}
\end{equation}
If a backbone contains an always-on shared expert $f_{\mathrm{sh}}$, the
complete MoE output may be written abstractly as
\begin{equation}
  y_{\mathrm{MoE}}(x)
  = y_{\mathrm{routed}}(x)
  + c_{\mathrm{sh}}(x)f_{\mathrm{sh}}(x),
  \label{eq:moe-shared}
\end{equation}
where $c_{\mathrm{sh}}(x)$ is the native shared-expert gate, or one for an
ungated shared expert.  This term is omitted for backbones without a shared
expert.  Router normalization, shared-expert gating, residual placement, and
normalization layers otherwise remain unchanged from the pretrained backbone.

Each routed expert contains fused gate/up and down projections.  We use
\emph{Weight Level} for directly adapting these expert weights,
\emph{Expert Level} for applying one adaptation branch to each routed expert,
and \emph{Module Level} for applying one shared adaptation branch to the
complete MoE module in a layer.  All unlisted pretrained parameters remain
frozen.

\paragraph{DoRA.}
DoRA~\citep{liu2024dora} separates the magnitude and direction of an adapted
weight.  Under the convention
$W_0\in\mathbb{R}^{d_{\mathrm{out}}\times d_{\mathrm{in}}}$ and $y=W_0x$,
define
\begin{align}
  V &= W_0 + \frac{\alpha}{r}BA,
  \\
  m_0 &= \lVert W_0\rVert_{\mathrm{row}},
  \qquad
  m\in\mathbb{R}^{d_{\mathrm{out}}},
\end{align}
where $m$ is trainable and initialized as $m_0$.  The adapted weight is
\begin{equation}
  W'
  = \operatorname{Diag}\!\left(
      m\oslash\lVert V\rVert_{\mathrm{row}}
    \right)V.
  \label{eq:dora}
\end{equation}
Here $\oslash$ denotes elementwise division.  This row-wise expression is the
transpose-equivalent form of the column-wise normalization used in the
original DoRA notation.

The memory-efficient implementation replaces the normalizer used for gradient
computation by
\begin{equation}
  c(V)
  = \operatorname{stopgrad}
    \!\left(\lVert V\rVert_{\mathrm{row}}\right),
  \qquad
  W'
  = \operatorname{Diag}\!\left(m\oslash c(V)\right)V.
  \label{eq:dora-detach}
\end{equation}
The stop-gradient operation changes the backward graph, not the numerical
forward value.  Weight-level DoRA is applied independently to the fused
gate/up and down projections of each adapted routed expert.

\paragraph{OFT.}
OFT~\citep{qiu2024oft} applies the same orthogonal transformation to the
weight vector of every output neuron in a frozen projection.  For a
skew-symmetric parameter $Q$,
\begin{equation}
  Q^\top=-Q,
  \qquad
  R=(I+Q)(I-Q)^{-1},
  \qquad
  R^\top R=I.
  \label{eq:oft}
\end{equation}
Under the convention $y=Wx$, with
$W\in\mathbb{R}^{d_{\mathrm{out}}\times d_{\mathrm{in}}}$, the adapted
projection is
\begin{equation}
  W'=WR^\top,
  \qquad
  y'=WR^\top x,
  \label{eq:oft-projection}
\end{equation}
Here $R\in\mathbb{R}^{d_{\mathrm{in}}\times d_{\mathrm{in}}}$ acts on the
input dimension, equivalently transforming each row weight vector by the
same orthogonal map.  For block-diagonal OFT,
\begin{equation}
  R=\operatorname{Diag}(R_1,\ldots,R_b),
\end{equation}
and the block size must divide the input width $d_{\mathrm{in}}$.  The
skew-symmetric parameters are initialized so that
$R=I$ at initialization.

\paragraph{FourierFT.}
FourierFT~\citep{gao2024fourierft} represents a dense weight update using a
small number of trainable Fourier coefficients.  Let $\Omega$ be a fixed
random support in the two-dimensional frequency domain, and let
$S_{\Omega}(c)$ scatter the trainable real coefficient vector $c$ into the
corresponding frequency locations.  The update is
\begin{equation}
  \Delta W
  = \gamma\,
    \operatorname{Re}\!\left[
      \operatorname{IDFT}_{2}\!\left(S_{\Omega}(c)\right)
    \right],
  \qquad
  W'=W+\Delta W.
  \label{eq:fourier}
\end{equation}
The original method samples the support once and uses randomly initialized
coefficients.  In our benchmark implementation, the support is fixed with
seed 2024, the coefficients are instead initialized to zero so that
$\Delta W=0$ initially, and the update scale is set to $\gamma=150$.
These are experimental implementation choices rather than defining
properties of FourierFT.

\paragraph{IA3.}
IA3~\citep{liu2022tfew} learns multiplicative activation vectors.  For routed
expert $e$, let
\begin{equation}
  h_e(x)
  =
  \phi\!\left(W_{\mathrm{gate},e}x\right)
  \odot W_{\mathrm{up},e}x.
\end{equation}
The feed-forward IA3 vector
$\ell_{\mathrm{ff},e}\in\mathbb{R}^{d_{\mathrm{ff}}}$ modifies the expert as
\begin{equation}
  f^{\mathrm{IA3}}_e(x)
  =
  W_{\mathrm{down},e}
  \left(\ell_{\mathrm{ff},e}\odot h_e(x)\right).
  \label{eq:ia3}
\end{equation}
All IA3 vectors are initialized to one.  Canonical IA3 additionally scales
attention key and value activations, but attention adaptation is disabled in
the final PEFT method comparison.

Where router IA3 is enabled in our benchmark, it is a MoE-specific extension
rather than part of canonical IA3.  It is applied before top-$k$ selection:
\begin{equation}
  g^{\mathrm{IA3}}(x)
  = \ell_{\mathrm{router}}\odot g(x),
  \qquad
  \ell_{\mathrm{router},0}=\mathbf{1},
  \label{eq:router-ia3}
\end{equation}
and $\mathcal{S}(x)$ and $p_e(x)$ are then computed from
$g^{\mathrm{IA3}}(x)$.

\paragraph{DAS-LoRA.}
DAS-LoRA~\citep{tang2026exploring} performs an offline analysis of native
router statistics on downstream-domain and reference data.  It computes the
paper's cumulative Domain Advantage Score (CDAS) for each routed expert and
selects a fixed budget of task-relevant experts in each layer:
\begin{equation}
  \mathcal{E}_{\ell}^{\mathrm{DAS}}
  =
  \operatorname{TopB}\!\left(
    \left\{\operatorname{CDAS}_{\ell,e}\right\}_{e=1}^{E},
    b_\ell
  \right),
  \label{eq:das-selection}
\end{equation}
where $b_\ell$ is the selected-expert budget.  We do not replace the paper's
CDAS procedure by a simple difference of mean routing probabilities.

LoRA is attached only to the gate/up/down weights of the selected experts.
Writing $\Delta f_{\ell,e}^{\mathrm{LoRA}}(x)$ for the change induced by
those weight-level low-rank updates, the layer output is
\begin{equation}
  y'_{\ell}(x)
  =
  \sum_{e\in\mathcal{S}_{\ell}(x)}
  p_{\ell,e}(x)
  \left[
    f_{\ell,e}(x)
    +
    \mathbb{I}
    \!\left(e\in\mathcal{E}_{\ell}^{\mathrm{DAS}}\right)
    \Delta f_{\ell,e}^{\mathrm{LoRA}}(x)
  \right].
  \label{eq:das-lora}
\end{equation}
Thus an adapted expert contributes only when it is both selected offline by
DAS-LoRA and dispatched to the current token by the native router.

\subsection{\texorpdfstring{MoE$^2$-LoRA}{MoE2-LoRA} and the Insight-Guided Redesign}
\label{app:moe2}

\paragraph{Canonical MoE$^2$-LoRA.}
MoE$^2$-LoRA~\citep{yang2026moe2lora} uses both native router information and a
low-dimensional hidden-state channel to route over a global pool of LoRA
experts.  For MoE layer $\ell$, let
\begin{align}
  z_\ell(x) &= g_\ell(x)\in\mathbb{R}^{E},\\
  h_{\ell}^{c}(x)
  &= W_{\ell}^{h}x\in\mathbb{R}^{d_b}.
\end{align}
The layer-specific LoRA-pool router is
\begin{equation}
  q_\ell(x)
  =
  \operatorname{softmax}
  \left(
    W_{\ell}^{q}
    \begin{bmatrix}
      z_\ell(x)\\
      h_{\ell}^{c}(x)
    \end{bmatrix}
  \right),
  \qquad
  W_{\ell}^{q}
  \in\mathbb{R}^{N_L\times(E+d_b)}.
  \label{eq:moe2-router}
\end{equation}
Let
\begin{equation}
  \mathcal{T}_{\ell}(x)
  =
  \operatorname{TopK}\!\left(q_\ell(x),K_L\right).
\end{equation}
The selected pool probabilities are renormalized as
\begin{equation}
  \widetilde q_{\ell,j}(x)
  =
  \frac{q_{\ell,j}(x)}
       {\sum_{i\in\mathcal{T}_{\ell}(x)}q_{\ell,i}(x)},
  \qquad
  j\in\mathcal{T}_{\ell}(x).
  \label{eq:moe2-normalize}
\end{equation}

A single LoRA expert pool
$\{(A_j,B_j)\}_{j=1}^{N_L}$ is shared across all MoE layers, while the routing
projections are layer-specific.  The canonical module-level update is
\begin{equation}
  y'_{\ell}(x)
  =
  y_{\mathrm{MoE},\ell}(x)
  +
  \sum_{j\in\mathcal{T}_{\ell}(x)}
  \widetilde q_{\ell,j}(x)
  \frac{\alpha}{r}B_jA_jx.
  \label{eq:moe2-forward}
\end{equation}
Our reproduced Qwen3 setting uses two layer groups with 128 rank-$32$ pool
entries each (256 total), top-$8$ routing, and a routing bottleneck of
$d_b=128$.

\paragraph{Weight-level redesign.}
Our redesign retains the native-router-conditioned pool selection but places
the resulting low-rank residuals directly in the expert weights rather than
applying a module-level adaptation after the complete MoE output.

Let
\begin{equation}
  \pi_\ell(j)=(e,t)
\end{equation}
map pool entry $j$ to routed expert $e$ and target projection
$t\in\{\mathrm{gu},\mathrm{down}\}$ in layer $\ell$.  For an input $u$ to that
projection, define the token-conditioned effective update
\begin{equation}
  \Delta W_{\ell,e,t}(x)
  =
  \sum_{j\in\mathcal{T}_{\ell}(x)}
  \mathbb{I}\!\left[\pi_\ell(j)=(e,t)\right]
  \widetilde q_{\ell,j}(x)
  \frac{\alpha_t}{r}B_jA_j.
  \label{eq:moe2-weight-update}
\end{equation}
The corresponding expert projection is evaluated as
\begin{equation}
  z'_{\ell,e,t}(u;x)
  =
  \left(
    W_{\ell,e,t}
    +
    \Delta W_{\ell,e,t}(x)
  \right)u.
  \label{eq:moe2-adapted-projection}
\end{equation}
Replacing the relevant projections in
Equations~\ref{eq:expert-gate-up}--\ref{eq:expert-mlp} yields the adapted
expert $\widehat f_{\ell,e}(x)$, and the redesigned routed output is
\begin{equation}
  y'_{\ell}(x)
  =
  \sum_{e\in\mathcal{S}_{\ell}(x)}
  p_{\ell,e}(x)\widehat f_{\ell,e}(x),
  \label{eq:moe2-redesign-output}
\end{equation}
with the backbone's shared-expert term added separately where applicable.

\section{Datasets, Evaluation, and Training Protocol}
\label{app:protocol}

\subsection{Downstream and General-Capability Evaluation}

The mathematical-reasoning experiments fine-tune on MetaMathQA-R1 (huggingface \nolinkurl{oumi-ai/MetaMathQA-R1}), a pre-constructed synthetic dataset whose problems are drawn from MetaMathQA~\citep{yu2024metamath} and whose responses are generated by DeepSeek-R1. We use only the final solution portion of each response for training, excluding the intermediate think content. Mathematical reasoning is evaluated on the 500 examples in MATH-500~\citep{hendrycks2021math}. The intent experiments fine-tune on ESFT\_Intent and report exact-match accuracy on a held-out set of 500 examples. We assess general-capability retention on MMLU-Pro~\citep{wang2024mmlupro} and IFEval~\citep{zhou2023ifeval}.

For each model--task--configuration in the four design studies, we train every candidate learning rate with three random seeds and select one rate by mean downstream score.  We then evaluate the three checkpoints at that rate on MMLU-Pro and IFEval.  Downstream and general-capability scores are averaged over the same three seeds; general scores do not affect learning-rate selection.  Table~\ref{tab:task-settings} summarizes the task-specific training and downstream-evaluation settings.

\begin{table}[t]
\centering
\small
\begin{tabular}{@{}lcc@{}}
\toprule
Setting & Math & Intent \\
\midrule
Training data & MetaMathQA-R1 & ESFT\_Intent \\
Maximum samples & 80,000 & 80,000 \\
Epochs & 1 & 3 \\
Sequence cutoff & 2,048 & 4,096 \\
Packing & default & disabled \\
Evaluation set size & 500 & 500 \\
Evaluation metric & accuracy & exact match \\
Maximum new tokens & 512 & 512 \\
Evaluation batch size & 32 & 8 \\
\bottomrule
\end{tabular}
\caption{Task-specific training and downstream-evaluation settings.}
\label{tab:task-settings}
\end{table}

\subsection{Common Optimization Settings}

All reported fine-tuning runs use four GPUs, BF16 computation, gradient checkpointing, and DeepSpeed ZeRO-2~\citep{rasley2020deepspeed}.  Optimization uses fused AdamW with $\beta_1=0.9$, $\beta_2=0.999$, and $\epsilon=10^{-8}$; the learning-rate schedule is cosine with a 0.1 warmup ratio.  For every experiment, the learning rate is selected by grid search over
candidates drawn from $\{10^{-5},10^{-4},10^{-3}\}$.  No adapter dropout was configured.  All backbones are loaded with remote model code enabled.

The per-device batch size is 6 with six accumulation steps on four GPUs, giving an effective batch size of $6\times6\times4=144$.  Gemma-4 Intent uses a per-device batch size of 4 and nine accumulation steps, preserving the same effective batch size of $4\times9\times4=144$.  GPT-OSS Intent explicitly disables FlashAttention.
\section{Experiment-Specific Settings}
\label{app:settings}

\subsection{Adaptation Position}
\label{app:position}

The position study uses one shared low-rank adaptation branch alongside the
complete expert block when experts are adapted, standard LoRA for attention,
and an optional LoRA branch on the native router.  All target layers are
enabled, pool and weight sharing are per layer,
and the auxiliary coefficient is zero.  The scale is $\alpha=2r$ throughout.
Each of the first three base configurations is evaluated both without and with
router LoRA ($r=8$, $\alpha=16$); the higher-budget configuration is evaluated
in the same two variants, yielding eight arms.  Table~\ref{tab:position-settings}
summarizes these adaptation-position configurations.

\begin{table}[t]
\centering
\small
\begin{tabular}{@{}lccc@{}}
\toprule
Base arm & Attention & Expert adaptation & Router variant \\
\midrule
Attention only & $r64/a128$ & -- & off or $r8/a16$ \\
Experts only & -- & $r256/a512$ & off or $r8/a16$ \\
Attention+experts & $r32/a64$ & $r128/a256$ & off or $r8/a16$ \\
Higher budget & $r64/a128$ & $r256/a512$ & off or $r8/a16$ \\
\bottomrule
\end{tabular}
\caption{Adaptation-position arms.  Expert adaptation uses one independently
routed top-1 branch per layer ($N_L=1$), shared by the complete expert block.}
\label{tab:position-settings}
\end{table}

For Qwen3, the estimated trainable counts before adding the optional router
branch are approximately 53.48M for attention $r=64$, 50.33M for expert
adaptation with $r=256$, 51.90M for attention $r=32$ plus expert adaptation
with $r=128$, and 103.81M for the
higher-budget arm.

\subsection{Adaptation Granularity}
\label{app:granularity-settings}

The final granularity experiment isolates the expert-adaptation design while retaining the
router LoRA anchor.  Attention is disabled.  Every arm targets all MoE
layers, uses per-layer pool/weight sharing, router LoRA $r=8$, $\alpha=16$,
and zero auxiliary coefficient.

\begin{table}[t]
\centering
\small
\begin{tabular}{@{}
  >{\raggedright\arraybackslash}p{0.29\columnwidth}
  >{\raggedright\arraybackslash}p{0.38\columnwidth}
  >{\raggedright\arraybackslash}p{0.22\columnwidth}@{}}
\toprule
Granularity & Routing and placement & Expert setting \\
\midrule
Module Level & Independent, $N_L=1$, top-1 & $r256/a512$ \\
\midrule
Expert Level & Follow native router; one branch per expert & $r2/a4$ \\
\midrule
Weight Level & Follow native router; gate/up/down weights & $r1/a2$ \\
\midrule
Weight Level & Follow native router; gate/up/down weights & $r2/a4$ \\
\bottomrule
\end{tabular}
\caption{The four final granularity arms.  Native routing uses 128 experts and
top-8 for Qwen3/Gemma-4, and 32 experts and top-4 for GPT-OSS.}
\label{tab:granularity-settings}
\end{table}

\subsection{PEFT Method Comparison}
\label{app:method-settings}

The ten-method comparison uses the same experts+router, no-attention protocol as the
final granularity study.  For LoRA-structured methods, the router anchor is
LoRA $r=8$, $\alpha=16$.  For DoRA, OFT, FourierFT, and IA3, the router uses
the same update family as the weight-level expert adaptation rather than a LoRA
adapter.  All methods target every MoE layer and use zero auxiliary loss.  The exact method-specific settings are listed in Tables~\ref{tab:method-settings-a}
and~\ref{tab:method-settings-b}.

\begin{table*}[t]
\centering
\small
\begin{tabular}{@{}
  >{\raggedright\arraybackslash}p{0.13\textwidth}
  >{\raggedright\arraybackslash}p{0.28\textwidth}
  >{\raggedright\arraybackslash}p{0.20\textwidth}
  >{\raggedright\arraybackslash}p{0.30\textwidth}@{}}
\toprule
Method & Expert update & Router update & Backbone-specific values \\
\midrule
Weight-level LoRA & Expert gate/up/down weights, $r1/a2$ & LoRA $r8/a16$ & Same on all three backbones \\
\midrule
Weight-level DoRA & Expert gate/up/down weights, $r1/a2$, plus magnitude & DoRA $r8/a16$ & Same ranks/scales on all three backbones \\
\midrule
Weight-level OFT & Blockwise Cayley transform & Blockwise OFT & Expert/router blocks: Qwen3 $4/128$; Gemma-4 $8/128$; GPT-OSS $6/32$ \\
\midrule
FourierFT & Fixed-support spectral delta & FourierFT & Expert/router frequencies: Qwen3 $4096/17408$; Gemma-4 $5632/23552$; GPT-OSS $5760/23296$; scale 150, seed 2024 \\
\midrule
IA3 & Per-expert down-input scale & Router-logit output scale & One-initialized vectors; no rank \\
\bottomrule
\end{tabular}
\caption{Weight-level methods in the final PEFT method comparison.  Attention
is disabled, so attention-specific capacity values are intentionally omitted.}
\label{tab:method-settings-a}
\end{table*}

\begin{table*}[t]
\centering
\small
\begin{tabular}{@{}
  >{\raggedright\arraybackslash}p{0.15\textwidth}
  >{\raggedright\arraybackslash}p{0.35\textwidth}
  >{\raggedright\arraybackslash}p{0.16\textwidth}
  >{\raggedright\arraybackslash}p{0.25\textwidth}@{}}
\toprule
Method & Expert update & Router update & Backbone-specific values \\
\midrule
Expert-level LoRA & One adaptation branch per routed expert, follows the native router's top-$k$, $r2/a4$ & LoRA $r8/a16$ & $E/k=128/8$ for Qwen3 and Gemma-4; $32/4$ for GPT-OSS \\
\midrule
MoE-LoRA & Independent adapter router, $N_L=8$, adapter top-2 & LoRA $r8/a16$ & FFN $r32/a64$ on Qwen3/Gemma-4; $r8/a16$ on GPT-OSS \\
\midrule
Module-level LoRA & One independent top-1 adaptation branch per layer & LoRA $r8/a16$ & FFN $r256/a512$ on Qwen3/Gemma-4; $r64/a128$ on GPT-OSS \\
\midrule
DAS-LoRA & Highest-DAS expert subset per layer & Enabled common LoRA router anchor & Select 8 experts with $r32/a64$ on Qwen3/Gemma-4; 4 experts with $r16/a32$ on GPT-OSS \\
\midrule
Partial FT & Full expert weights in last decoder layer & Full router in last layer & No adapter and no parameter matching \\
\bottomrule
\end{tabular}
\caption{MoE-structured and partial-fine-tuning methods in the final PEFT
method comparison.  MoE$^2$-LoRA is excluded from this controlled
ten-method grid and analyzed separately in Section~\ref{app:moe2}.}
\label{tab:method-settings-b}
\end{table*}

\subsection{Auxiliary Load-Balancing Loss}
\label{app:aux-settings}

The auxiliary study uses weight-level LoRA with $r=1$, $\alpha=2$,
router LoRA $r=8$, $\alpha=16$, no attention, and the same learning-rate grid
as the granularity study.  For a batch-level routing frequency $f_e$ and mean
dense router probability $\bar p_e$, the per-layer balancing term has the form
\begin{equation}
  \mathcal{L}_{\mathrm{bal}} = E\sum_{e=1}^{E} f_e\bar p_e,
  \qquad
  \mathcal{L}=\mathcal{L}_{\mathrm{task}}+\lambda\mathcal{L}_{\mathrm{bal}}.
  \label{eq:balance-loss}
\end{equation}
The schedule is constant and
$\lambda\in\{0,10^{-4},10^{-3},10^{-2},10^{-1}\}$.  The zero-coefficient arm
reuses the corresponding no-attention granularity baseline.  The downstream
averages reported in the main paper are 68.5, 68.1, 66.8, 64.6, and 58.6
percent for increasing $\lambda$, respectively.

\section{Additional \texorpdfstring{MoE$^2$-LoRA}{MoE2-LoRA} Validation of Placement and Balancing}
\label{app:moe2-validation}

In this section, we test the adaptation-position conclusion (Insight~1 in the main paper) and the auxiliary-load-balancing conclusion (Insight~4).

\subsection{Experimental Controls}

The experts-only $r32$ reference uses the reproduced MoE$^2$-LoRA
configuration: two layer groups, a 128-entry rank-32 adaptation pool per group
(256 entries in total),
$\alpha=64$, learned top-8 pool routing, a 128-dimensional routing bottleneck,
and no attention or router LoRA.  Its load-balancing coefficient is zero.  The
budget-matched placement arm reduces the expert-adaptation pool to $r=18$,
$\alpha=36$, and
adds standard rank-18, $\alpha=36$ LoRA to the Q, K, V, and output projections
in every attention layer.  This changes the trainable count by only 0.76\%,
from 47.71M to 48.07M parameters.

The balancing arms keep the experts-only $r32$ architecture and trainable count fixed,
leave attention and router adaptation disabled, and apply a constant
$\lambda\in\{10^{-4},10^{-3},10^{-2},10^{-1}\}$.  The downstream-selected
learning rate is $10^{-3}$ for Intent and $10^{-4}$ for Math.  Intent is
trained for three epochs and Math for one epoch; both use the common
80,000-sample training and downstream-selected checkpoint protocol.  MMLU-Pro
and IFEval are evaluated on those same checkpoints.
Tables~\ref{tab:moe2-validation-intent}
and~\ref{tab:moe2-validation-math} report full-set results.

\begin{table*}[t]
\centering
\small
\setlength{\tabcolsep}{8pt}
\begin{tabular}{@{}lrrrr@{}}
\toprule
Configuration & Trainable params & Intent & MMLU-Pro & IFEval \\
\midrule
Experts only
  & 47.71M & 84.00 & 45.26 & 18.11 \\
Attention+experts
  & 48.07M & 81.60 & 47.93 & 14.23 \\
Balance $10^{-4}$
  & 47.71M & 85.00 & 47.46 & 17.56 \\
Balance $10^{-3}$
  & 47.71M & 82.20 & 45.18 & 5.91 \\
Balance $10^{-2}$
  & 47.71M & 80.80 & 22.37 & 7.76 \\
Balance $10^{-1}$
  & 47.71M & 67.80 & 16.89 & 13.31 \\
\bottomrule
\end{tabular}
\caption{Additional Qwen3 MoE$^2$-LoRA validation on Intent.  All metrics are
percentages.  Intent uses 500 held-out examples, MMLU-Pro uses 12,032 examples,
and IFEval is prompt-level accuracy on 541 prompts.}
\label{tab:moe2-validation-intent}
\end{table*}

\begin{table*}[t]
\centering
\small
\setlength{\tabcolsep}{7pt}
\begin{tabular}{@{}lrrrr@{}}
\toprule
Configuration & Trainable params & MATH-500 & MMLU-Pro & IFEval \\
\midrule
Experts only
  & 47.71M & 60.00 & 48.19 & 35.12 \\
Attention+experts
  & 48.07M & 56.60 & 47.56 & 36.23 \\
Balance $10^{-4}$
  & 47.71M & 56.40 & 47.27 & 40.67 \\
Balance $10^{-3}$
  & 47.71M & 55.80 & 46.69 & 43.81 \\
Balance $10^{-2}$
  & 47.71M & 55.60 & 44.83 & 45.47 \\
Balance $10^{-1}$
  & 47.71M & 53.40 & 42.50 & 46.58 \\
\bottomrule
\end{tabular}
\caption{Additional Qwen3 MoE$^2$-LoRA validation on Math.  All metrics are
percentages.}
\label{tab:moe2-validation-math}
\end{table*}

\subsection{Evidence for the Two Design Defaults}

\paragraph{Adaptation position.}
Starting from the experts-only reference, adding attention adaptation at an
effectively matched parameter budget consistently weakens downstream
performance: Intent decreases from 84.00 to 81.60 and MATH-500 from 60.00 to
56.60.  For each task, the joint Attention+experts configuration also
degrades at least one upstream benchmark.  On Intent, IFEval falls from 18.11 to 14.23
despite the MMLU-Pro improvement; on Math, MMLU-Pro falls from 48.19 to 47.56
despite the IFEval improvement.  Adding attention therefore fails to provide a
consistent benefit and instead introduces simultaneous downstream and upstream
degradation on both tasks.  These results reinforce Insight~1: for MoE
fine-tuning, target the experts by default and unfreeze shared attention
modules cautiously, since adapting more modules does not necessarily yield
better transfer and may amplify general-capability forgetting.

\paragraph{Auxiliary load balancing.}
On Intent, the smallest coefficient, $10^{-4}$, remains competitive: it
improves Intent from 84.00 to 85.00 and MMLU-Pro from 45.26 to 47.46, with only
a 0.55-point decrease on IFEval.  This local benefit does not transfer to Math,
where the same coefficient reduces MATH-500 from 60.00 to 56.40 and MMLU-Pro
from 48.19 to 47.27, although IFEval improves.  The other nonzero coefficients
are generally more damaging.  They all lower downstream performance on both
tasks; on Intent, the larger coefficients also cause substantial upstream
degradation, while on Math the IFEval gains are consistently accompanied by
losses on both MATH-500 and MMLU-Pro.  Thus, auxiliary balancing does not yield
a task-stable improvement.  These results support Insight~4: the auxiliary
loss should be disabled or assigned a very small coefficient during MoE
adaptation.  A coefficient around $10^{-4}$ is a safe choice when
routing regularization is required and may benefit particular datasets,
whereas larger values can substantially degrade downstream performance.

\section{Backbone Architectures}
\label{app:backbones}

The experiments use the text-generation pathways of
Qwen3-30B-A3B-Base~\citep{qwen2025qwen3},
Gemma-4-26B-A4B~\citep{gemmateam2026gemma4technicalreport}, and
GPT-OSS-20B~\citep{openai2025gptoss20b}.  Gemma-4 is distributed through a
multimodal wrapper, but no visual inputs or vision-tower adaptation are used.
The architectural values below follow the official reports, model cards, and
released checkpoint configurations.  Tables~\ref{tab:attention-architecture}
and~\ref{tab:moe-architecture} summarize the attention and MoE architectures,
respectively.

\begin{table*}[t]
\centering
\small
\setlength{\tabcolsep}{4pt}
\begin{tabular}{@{}>{\raggedright\arraybackslash}p{0.17\textwidth}
  >{\centering\arraybackslash}p{0.065\textwidth}
  >{\centering\arraybackslash}p{0.07\textwidth}
  >{\centering\arraybackslash}p{0.065\textwidth}
  >{\centering\arraybackslash}p{0.105\textwidth}
  >{\centering\arraybackslash}p{0.105\textwidth}
  >{\raggedright\arraybackslash}p{0.295\textwidth}@{}}
\toprule
Backbone & Layers & Hidden & Q heads & KV heads & Head dim & Attention pattern \\
\midrule
Qwen3-30B-A3B-Base & 48 & 2048 & 32 & 4 & 128 & Full attention; no sliding window \\
\midrule
Gemma-4-26B-A4B & 30 & 2816 & 16 & 8 local / 2 global & 256 local / 512 global & Five sliding layers followed by one global layer; window 1024; global K=V and p-RoPE \\
\midrule
GPT-OSS-20B & 24 & 2880 & 64 & 8 & 64 & Alternating sliding and full layers; window 128 \\
\bottomrule
\end{tabular}
\caption{Text-backbone and attention differences.  ``Q heads'' and ``KV
heads'' distinguish query heads from grouped key/value heads.  Gemma-4 changes
both KV-head count and head dimension in its global-attention layers and ties
their key and value projections.}
\label{tab:attention-architecture}
\end{table*}

\begin{table*}[t]
\centering
\small
\setlength{\tabcolsep}{4pt}
\begin{tabular}{@{}>{\raggedright\arraybackslash}p{0.17\textwidth}
  >{\centering\arraybackslash}p{0.08\textwidth}
  >{\centering\arraybackslash}p{0.055\textwidth}
  >{\raggedright\arraybackslash}p{0.235\textwidth}
  >{\centering\arraybackslash}p{0.08\textwidth}
  >{\raggedright\arraybackslash}p{0.25\textwidth}@{}}
\toprule
Backbone & Routed experts & Top-$k$ & Shared/always-on expert path & Routed width & Expert adaptation target \\
\midrule
Qwen3-30B-A3B-Base & 128 & 8 & No explicit shared expert & 768 & Routed \texttt{experts} host \\
\midrule
Gemma-4-26B-A4B & 128 & 8 & One always-active shared expert; dense width 2112 in config & 704 & The 128 routed \texttt{experts}; shared path frozen \\
\midrule
GPT-OSS-20B & 32 & 4 & No explicit shared expert & 2880 & Routed \texttt{experts} host \\
\bottomrule
\end{tabular}
\caption{MoE structure and the actual expert-adaptation boundary.  ``No explicit
shared expert'' means that no such branch is defined in the released
checkpoint configuration or reference implementation; it is not a claim about
all models in that family.  The Gemma-4 shared path is always active, while the
repository adapts only its separate routed-expert container.}
\label{tab:moe-architecture}
\end{table*}

These backbones are not minor variants of one MoE template.  They
differ in expert cardinality (32 versus 128), token-level fan-out (top-4 versus
top-8), routed width (704, 768, or 2880), the presence of an always-active
shared expert, grouped-KV geometry, and attention locality.  In particular,
Gemma-4 tests routed-expert adaptation in the presence of a frozen shared path,
while Qwen3 and GPT-OSS expose no explicit shared-expert branch; Qwen3 uses
full attention throughout, whereas the other two mix local and global
attention with different schedules.  Against this architectural variation,
the recurring trends behind the main paper's placement, granularity, PEFT
method, and balancing insights are observed across the three-backbone
grid, even though individual endpoints need not share an identical ordering.
This makes the findings less likely to be an artifact of one expert count,
router fan-out, or attention topology, while remaining empirical rather than
universal claims.

\section{Functional Geometry of Adaptation Granularity}
\label{app:functional-rank}

This appendix gives the response definitions, rank derivations, measurement
protocol, and detailed numerical results for the functional analysis of
adaptation granularity. The module-level rank constrains one shared branch,
the expert-level rank constrains each routed branch, and the weight-level rank
constrains individual expert projections. We compare these configurations
through the output of the full MoE computation, including nonlinear expert
features and routed aggregation.

\subsection{Layer-Level Response and Functional Effective Rank}
\label{app:functional-rank-definition}

For MoE layer $\ell$, let $F_{0,\ell}(x)$ denote the frozen base-model block
output and $F_{a,\ell}(x)$ the output after loading adapter configuration $a$.
The adapter-induced functional perturbation is
\begin{equation}
  \delta F_{a,\ell}(x)
  = F_{a,\ell}(x)-F_{0,\ell}(x).
  \label{eq:functional-perturbation}
\end{equation}
For a probe set $\mathcal{X}_{\ell}=\{x_1,\ldots,x_n\}$, we collect these
perturbations as columns,
\begin{equation}
  Y_{a,\ell}
  = \begin{bmatrix}
      \delta F_{a,\ell}(x_1) & \cdots & \delta F_{a,\ell}(x_n)
    \end{bmatrix}
  \in\mathbb{R}^{d\times n}.
  \label{eq:functional-response-matrix}
\end{equation}
Unlike the rank of an individual LoRA factor or weight matrix,
$Y_{a,\ell}$ includes the effect of pretrained expert computation, routing
weights, and aggregation across selected experts.

We define the uncentered second-moment matrix
\begin{equation}
  K_{a,\ell}=\frac{1}{n}Y_{a,\ell}Y_{a,\ell}^{\top}.
  \label{eq:functional-second-moment}
\end{equation}
If its eigenvalues are $\lambda_1,\ldots,\lambda_d\geq 0$, the layer-level
functional effective rank is the participation ratio
\begin{equation}
  r_{\mathrm{eff}}(Y_{a,\ell})
  = \frac{\bigl(\operatorname{tr}K_{a,\ell}\bigr)^2}
         {\lVert K_{a,\ell}\rVert_F^2+\varepsilon}
  = \frac{\bigl(\sum_i\lambda_i\bigr)^2}
         {\sum_i\lambda_i^2+\varepsilon},
  \label{eq:functional-effective-rank}
\end{equation}
where $\varepsilon$ is used only for numerical protection.  For a nonzero
response in the $\varepsilon\rightarrow 0$ limit, concentrating all energy in
one direction gives $r_{\mathrm{eff}}=1$, whereas equal energy in $k$
orthogonal directions gives $r_{\mathrm{eff}}=k$.  More generally,
\begin{equation}
  r_{\mathrm{eff}}(Y_{a,\ell})
  \leq \operatorname{rank}(Y_{a,\ell})
  \leq \min(d,n).
\end{equation}
The measure is invariant to a common nonzero rescaling of the response, up to
the numerical stabilizer.  It therefore describes how response energy is
distributed across directions, rather than the perturbation magnitude.  A
larger value indicates a higher-dimensional realized response on the probe
set. Table~\ref{tab:functional-effective-rank} reports this geometric
quantity alongside the measured downstream and retention outcomes.

\subsection{Why Local Rank Does Not Determine Layer-Level Rank}
\label{app:functional-rank-mechanism}

\paragraph{Module- and expert-level adaptation.}
A module-level adaptation produces the linear correction
\begin{equation}
  \delta F_{M}(x)=Cx,
  \qquad \operatorname{rank}(C)\leq r_M.
  \label{eq:module-level-response}
\end{equation}
Writing the probe inputs as $X=[x_1,\ldots,x_n]$ gives
\begin{equation}
  \operatorname{rank}(Y_M)
  =\operatorname{rank}(CX)
  \leq \min\bigl(r_M,\operatorname{rank}(X)\bigr).
  \label{eq:module-level-rank-bound}
\end{equation}
Thus, $r_M=256$ is an upper bound for the single shared map, not its realized
effective rank.  A concentrated learned spectrum or a restricted probe span
can yield a substantially smaller value.  Moreover, this local bound cannot
be converted directly into the rank of a configuration distributed across
many experts and projections.

For expert-level adaptation, the correction from expert $e$ is
\begin{equation}
  \delta f_{E,e}(x)=C_ex,
  \qquad \operatorname{rank}(C_e)\leq r_E.
  \label{eq:expert-level-response}
\end{equation}
Multiplication by the input-dependent routing weight $p_e(x)$ does not move a
branch response outside $\operatorname{col}(C_e)$.  The complete layer,
however, aggregates several selected experts whose column spaces may differ;
its effective rank can therefore exceed $r_E$.  The value $r_E=2$ constrains
each branch, not the routed layer response.

\paragraph{Weight-level adaptation.}
Write one gated expert as
\begin{equation}
  f_e(x)=D_e\bigl[\sigma(G_ex)\odot U_ex\bigr],
  \label{eq:gated-expert}
\end{equation}
where $G_e$, $U_e$, and $D_e$ are the gate, up, and down projections.  Consider
only a rank-one update to the up projection,
$\Delta U_e=ab^{\top}$.  Its exact output correction is
\begin{equation}
  \delta_U f_e(x)
  =(b^{\top}x)D_e\operatorname{Diag}(a)\sigma(G_ex).
  \label{eq:rank-one-up-response}
\end{equation}
For $n$ inputs, define
\begin{equation}
  S_{\mathcal{X}}
  =\begin{bmatrix}
      \sigma(G_ex_1)&\cdots&\sigma(G_ex_n)
    \end{bmatrix},
  \qquad \zeta_j=b^{\top}x_j.
\end{equation}
The resulting sample-response matrix is
\begin{equation}
  \begin{aligned}
    Y_U
    &=D_e\operatorname{Diag}(a)S_{\mathcal{X}}
      \operatorname{Diag}(\zeta), \\
    \operatorname{rank}(Y_U)
    &\leq\min(d,m,n),
  \end{aligned}
  \label{eq:rank-one-up-response-matrix}
\end{equation}
where $m$ is the expert intermediate width.  When
$D_e\operatorname{Diag}(a)S_{\mathcal{X}}$ attains the corresponding rank and
all $\zeta_j$ are nonzero, this upper bound is attainable.  Hence, even a
rank-one update to a single expert projection can reuse the frozen gate's
input-dependent nonlinear features to produce a multi-directional sample
response.  The measured weight-level configurations update gate, up, and down
projections across all 128 experts and then aggregate routed responses.
Equation~\ref{eq:rank-one-up-response-matrix} establishes the rank-expansion
mechanism for an individual up-projection update. The full measurements
include all adapted projections and routed aggregation.

\subsection{Measurement Protocol}
\label{app:functional-rank-protocol}

We measure the four Qwen3 granularity configurations selected for the Math and
Intent experiments.  The protocol is as follows.
\begin{itemize}
  \item We scan all 48 MoE layers and all 128 routed experts, without expert
  subsampling.
  \item Each layer uses $n=384$ deterministic synthetic probes,
  $x\sim\mathcal{N}\bigl(0,\operatorname{diag}(\gamma_{\ell}^{2})\bigr)$,
  where $\gamma_{\ell}$ is the pretrained pre-MoE RMSNorm gain for that layer.
  \item We estimate a checkpoint-specific expert-selection distribution from
  the recorded training-time activation counts.  For each synthetic token, we
  sample eight experts without replacement and assign each selected expert an
  aggregation weight of $1/8$.
  \item The expert-level and two weight-level configurations share the same
  synthetic routing realization.  The module-level adaptation is applied once
  per token and does not pass through expert routing.
  \item We evaluate Equation~\ref{eq:functional-effective-rank} for each
  task--configuration--layer combination, giving
  $2\times4\times48=384$ layer-level observations.
\end{itemize}
Math and Intent identify the trained adapters and their recorded routing
statistics. All reported functional ranks characterize these checkpoints on
the RMSNorm-scaled synthetic probe distribution specified above.

\subsection{Functional Effective Rank Across Granularities}
\label{app:functional-rank-results}

\begin{table*}[t]
\centering
\small
\setlength{\tabcolsep}{5pt}
\begin{tabular}{@{}lcrrrrp{0.28\textwidth}@{}}
\toprule
Granularity & Local $r$ & Math & Intent & Mean & Down. avg. & General-capability observation \\
\midrule
Module Level
  & 256 & 25.25 & 20.54 & 22.90 & 68.4
  & MMLU-Pro $-0.5$; IFEval $-3.7$ \\
\midrule
Expert Level
  & 2 & 16.31 & 15.91 & 16.11 & 64.7
  & MMLU-Pro $-5.0$; IFEval $-4.6$; unstable across backbones \\
\midrule
Weight Level
  & 1 & 63.90 & 69.08 & 66.49 & 68.5
  & MMLU-Pro $+0.8$; IFEval $-1.4$; best retention \\
\midrule
Weight Level
  & 2 & \textbf{86.37} & \textbf{85.22} & \textbf{85.80} & \textbf{69.1}
  & MMLU-Pro $-0.9$; IFEval $-3.4$; nearly twice the trainable parameters \\
\bottomrule
\end{tabular}
\caption{Functional effective rank and the corresponding granularity results.
Math, Intent, and Mean report Qwen3 layer-level effective ranks; Mean averages
the 96 task--layer observations.  Downstream and general-capability results
are the main-paper aggregates over three backbones and two task endpoints.
The local $r$ values constrain different objects and are included only to
identify each configuration.}
\label{tab:functional-effective-rank}
\end{table*}

Table~\ref{tab:functional-effective-rank} reports the checkpoint-specific
means and their average over 96 task--layer observations. Each layer uses
384 probes, so the response matrix has rank at most 384. The reported
participation ratios further account for the concentration of energy within
that response span.

\subsection{Configuration-Level Comparisons}
\label{app:functional-rank-relation}

\paragraph{Within weight-level adaptation.}
Holding the expert and projection set fixed, increasing $r$ from one to two
raises the mean effective rank from 66.49 to 85.80, a 29.0\% increase.
The task-specific increases are 22.47 on Math (35.2\%) and 16.14 on Intent
(23.4\%). The corresponding three-backbone downstream average rises from
68.5 to 69.1, while rank one uses approximately half the trainable parameters
and retains more general capability. These measurements identify the
additional functional coverage supplied by a larger local rank and its
associated task, parameter, and retention costs.

\paragraph{Across adaptation positions.}
The expert-level responses have mean effective ranks near 16 on both
checkpoints. Their fixed branch column spaces are combined through routing,
as characterized in Appendix~\ref{app:functional-rank-mechanism}.
The shared module-level branch reaches 25.25 on Math and 20.54 on Intent,
showing different energy concentration under the same local rank of 256.
Weight-level rank one reaches 63.90 and 69.08 through the nonlinear mechanism
in Equation~\ref{eq:rank-one-up-response-matrix}.

\paragraph{Interpreting the reported quantities.}
Functional ranks are measured on the selected Qwen3 checkpoints, while the
downstream and retention columns summarize the three-backbone study.
Effective rank characterizes response diversity; MMLU-Pro and IFEval directly
measure general-capability retention. Together with the expert-local spectral
measurements below, these quantities describe complementary aspects of the
adaptation trade-off.

\FloatBarrier

\section{Pretrained Spectral Consistency Across Adaptation Granularities}
\label{app:spectral-consistency}

This appendix provides the expert-local spectral definitions, the mechanism
of spectral inheritance, and the full measurement protocol and numerical
profiles. We express each expert-output perturbation in the left singular
basis of its corresponding pretrained down projection and compare its
cumulative energy distribution with that projection's spectrum.

\subsection{Spectral Profile and Consistency Score}
\label{app:spectral-consistency-definition}

Let the pretrained down projection of expert $e$ be
$D_e\in\mathbb{R}^{d\times m}$, with singular value decomposition
\begin{equation}
  \begin{aligned}
    D_e
    &=U_e\Sigma_eV_e^{\top} \\
    &=\sum_{i=1}^{m}\sigma_{e,i}u_{e,i}v_{e,i}^{\top}.
  \end{aligned}
  \label{eq:pretrained-down-svd}
\end{equation}
The left singular vectors $u_{e,i}$ form the pretrained expert's output
coordinates, with
$\sigma_{e,1}\geq\cdots\geq\sigma_{e,m}\geq0$, ordered from spectrally strong
to weak directions.  For adapter
configuration $a$ and probes $\mathcal{X}=\{x_1,\ldots,x_n\}$, define the
expert-output perturbation matrix
\begin{equation}
  \begin{aligned}
    Y_{a,e}
    &=\begin{bmatrix}
       \delta f_{a,e}(x_1)&\cdots&\delta f_{a,e}(x_n)
      \end{bmatrix}, \\
    \delta f_{a,e}(x)
    &=f_{a,e}(x)-f_{0,e}(x).
  \end{aligned}
  \label{eq:expert-spectral-response}
\end{equation}
For weight-level adaptation, this is the change induced jointly by the gate,
up, and down LoRA updates.  For expert-level adaptation, it is the output of
the new expert-local branch.  Measuring experts separately ensures that every
response is compared with its corresponding pretrained $D_e$.

The cumulative fraction of perturbation energy carried by the first $k$
pretrained output directions is
\begin{equation}
  S_k(Y_{a,e})
  =\frac{
      \sum_{i\leq k}\lVert u_{e,i}^{\top}Y_{a,e}\rVert_2^2
    }{
      \sum_{i\leq m}\lVert u_{e,i}^{\top}Y_{a,e}\rVert_2^2
      +\varepsilon
    }.
  \label{eq:cumulative-update-spectrum}
\end{equation}
We compare this profile with the down projection's own cumulative spectrum,
\begin{equation}
  S_{k,e}^{0}
  =\frac{\sum_{i\leq k}\sigma_{e,i}^{2}}
         {\sum_{i\leq m}\sigma_{e,i}^{2}},
  \qquad
  S_k^{\mathrm{uni}}=\frac{k}{m},
  \label{eq:pretrained-uniform-spectral-references}
\end{equation}
where $S_k^{\mathrm{uni}}$ is the reference for energy distributed uniformly
over the $m$ pretrained directions.  A profile close to $S_{k,e}^{0}$ follows
the pretrained strong-to-weak hierarchy, whereas a profile close to $k/m$ is
approximately uniform in the same singular coordinates.

To summarize the three measured cutoffs
$\mathcal{K}=\{16,64,256\}$, we define the pretrained spectral-consistency
score
\begin{equation}
  C_{\mathrm{spec}}(Y_{a,e})
  =1-
  \frac{
    \sum_{k\in\mathcal{K}}
    \left|S_k(Y_{a,e})-S_{k,e}^{0}\right|
  }{
    \sum_{k\in\mathcal{K}}
    \left|k/m-S_{k,e}^{0}\right|+\varepsilon
  }.
  \label{eq:spectral-consistency}
\end{equation}
This normalization assigns $C_{\mathrm{spec}}=1$ to an exact match with the
pretrained cumulative profile and $C_{\mathrm{spec}}=0$ to the uniform
reference; negative values indicate a larger aggregate deviation from the
pretrained profile than the uniform reference.  Because common rescaling of
$Y_{a,e}$ leaves the score unchanged, it isolates the relative spectral
organization of the realized perturbation.

The pretrained expert computation provides a direct mechanism for spectral
inheritance.
For a perturbation component that passes through the frozen down projection,
$Y_{a,e}^{\parallel}=D_eH_{a,e}$, Equation~\ref{eq:pretrained-down-svd} gives
\begin{equation}
  S_k(Y_{a,e}^{\parallel})
  =\frac{
    \sum_{i\leq k}\sigma_{e,i}^{2}
    \lVert v_{e,i}^{\top}H_{a,e}\rVert_2^2
  }{
    \sum_{i\leq m}\sigma_{e,i}^{2}
    \lVert v_{e,i}^{\top}H_{a,e}\rVert_2^2+\varepsilon
  }.
  \label{eq:conditional-spectral-inheritance}
\end{equation}
Thus, when the latent perturbation energy is broadly distributed in the right
singular coordinates, the pretrained singular values shape the output-energy
profile.  Weight-level gate/up updates traverse this frozen projection and can
inherit its hierarchy.  By contrast, expert-level adaptation produces its
output through a newly learned map, without the same architectural coupling to the
expert's pretrained down spectrum.  The measured score tests the resulting
placement-dependent distinction directly.

\subsection{Measurement Protocol}
\label{app:spectral-consistency-protocol}

We evaluate the Qwen3 weight-level $r=1$, weight-level $r=2$, and expert-level
$r=2$ checkpoints selected for the Math and Intent experiments.
The protocol is as follows.
\begin{itemize}
  \item We scan all 48 MoE layers, $\ell\in\{0,\ldots,47\}$, and all 128
  routed experts in each layer, giving 6,144 layer--expert positions per
  checkpoint and configuration.
  \item Each layer uses $n=768$ deterministic Gaussian probes scaled by the
  pretrained pre-MoE RMSNorm gain.
  \item We compute the expert-output perturbation before cross-expert
  aggregation and express it in the left singular basis of the corresponding
  pretrained down projection.
  \item For Qwen3, $m=768$; we record $S_{16}$, $S_{64}$, and $S_{256}$ and
  evaluate Equation~\ref{eq:spectral-consistency} at every layer--expert
  position before averaging the scores.
\end{itemize}
The diagnostic is expert-indexed.  The module-level configuration produces
one shared post-MoE response rather than an expert-specific response, and its
geometry is
therefore characterized by the layer-level analysis in
Section~\ref{app:functional-rank}.

\subsection{Spectral Consistency Across Granularities}
\label{app:spectral-consistency-results}

\begin{table*}[t]
\centering
\small
\setlength{\tabcolsep}{7pt}
\begin{tabular}{@{}llrrrr@{}}
\toprule
Checkpoint & Configuration & $S_{16}$ & $S_{64}$ & $S_{256}$ & $C_{\mathrm{spec}}$ \\
\midrule
Reference & Pretrained down spectrum & 0.09368 & 0.25758 & 0.64587 & 1.000 \\
Reference & Uniform directions       & 0.02083 & 0.08333 & 0.33333 & 0.000 \\
\midrule
Math & Weight Level, $r=1$ & 0.08921 & 0.23353 & 0.58366 & \textbf{0.812} \\
Math & Weight Level, $r=2$ & 0.08950 & 0.23470 & 0.58642 & \textbf{0.822} \\
Math & Expert Level, $r=2$ & 0.03762 & 0.11561 & 0.37202 & 0.141 \\
\midrule
Intent & Weight Level, $r=1$ & 0.08308 & 0.21957 & 0.55615 & \textbf{0.733} \\
Intent & Weight Level, $r=2$ & 0.08338 & 0.22072 & 0.55822 & \textbf{0.741} \\
Intent & Expert Level, $r=2$ & 0.03151 & 0.10275 & 0.35596 & 0.087 \\
\bottomrule
\end{tabular}
\caption{Mean cumulative spectral profiles and pretrained
spectral-consistency scores.  Each adapted row averages 6,144 layer--expert
positions.  The reference rows average the corresponding pretrained expert
spectra and use $m=768$ for the uniform profile.}
\label{tab:spectral-consistency}
\end{table*}

Table~\ref{tab:spectral-consistency} reports the three cumulative energy
cutoffs and the mean consistency score. Scores are computed separately for
each layer--expert position before averaging, preserving each expert's own
pretrained reference. Weight-level rank one exceeds expert-level adaptation
by 0.671 on Math and 0.646 on Intent in mean $C_{\mathrm{spec}}$.
The layer-mean difference is positive in all 48 layers on both checkpoints;
paired differences are positive at 99.8\% of Math positions and 96.3\% of
Intent positions. The two weight-level ranks produce closely matched
cumulative profiles at all three cutoffs.

\subsection{Alignment with the Main Granularity Result}
\label{app:spectral-consistency-relation}

\begin{table}[t]
\centering
\footnotesize
\setlength{\tabcolsep}{2.5pt}
\begin{tabular}{@{}lrrrr@{}}
\toprule
Configuration & $C_{\mathrm{spec}}$ & Down. & MMLU & IFEval \\
              & (M / I) & avg. & $\Delta$ & $\Delta$ \\
\midrule
Weight Level $r=1$ & \textbf{0.812 / 0.733} & 68.5 & \textbf{$+0.8$} & \textbf{$-1.4$} \\
Weight Level $r=2$ & \textbf{0.822 / 0.741} & \textbf{69.1} & $-0.9$ & $-3.4$ \\
Expert Level $r=2$ & 0.141 / 0.087 & 64.7 & $-5.0$ & $-4.6$ \\
\bottomrule
\end{tabular}
\caption{Pretrained spectral consistency alongside the main granularity
results.  M/I denotes Math/Intent, and MMLU denotes MMLU-Pro.  Spectral scores
are measured on Qwen3; downstream and general-capability values are the
main-paper aggregates over three backbones and two task endpoints.}
\label{tab:spectral-consistency-relation}
\end{table}

Table~\ref{tab:spectral-consistency-relation} places the Qwen3 spectral
measurements beside the three-backbone granularity results. Both
weight-level configurations combine high spectral consistency with better
downstream accuracy and retention than the expert-level configuration.
Within weight-level adaptation, rank two increases functional response
diversity while rank one gives the best retention and near-best downstream
accuracy. The detailed measurements thus connect update placement to both
response diversity and the reuse of pretrained spectral structure.

\FloatBarrier

\subsection{Energy in the Pretrained Output Subspace}
\label{app:subspace-energy}

Weight-level adaptation also concentrates more perturbation energy in the
pretrained expert's output subspace.  To quantify this concentration, we
measure the fraction of the expert-output perturbation carried by
$\operatorname{col}(D_e)$:
\begin{equation}
  \eta_{\mathrm{col}}(Y_{a,e})
  =\frac{\lVert U_e^{\top}Y_{a,e}\rVert_F^2}
         {\lVert Y_{a,e}\rVert_F^2+\varepsilon}.
  \label{eq:subspace-energy}
\end{equation}
Here $U_e$ contains the $m=768$ left singular directions of the pretrained
down projection.  We measure this share on RMSNorm-scaled Gaussian probes
across all 48 layers and 128 experts for each Math and Intent checkpoint.
The projection energy is estimated using
$D_e(D_e^{\top}D_e+\lambda_e I)^{-1}D_e^{\top}$, with
$\lambda_e=10^{-6}\operatorname{tr}(D_e^{\top}D_e)/m$.
An isotropic output perturbation has expected share $m/d=768/2048=0.375$.

\begin{figure}[!ht]
\centering
\includegraphics[width=1.0\linewidth]{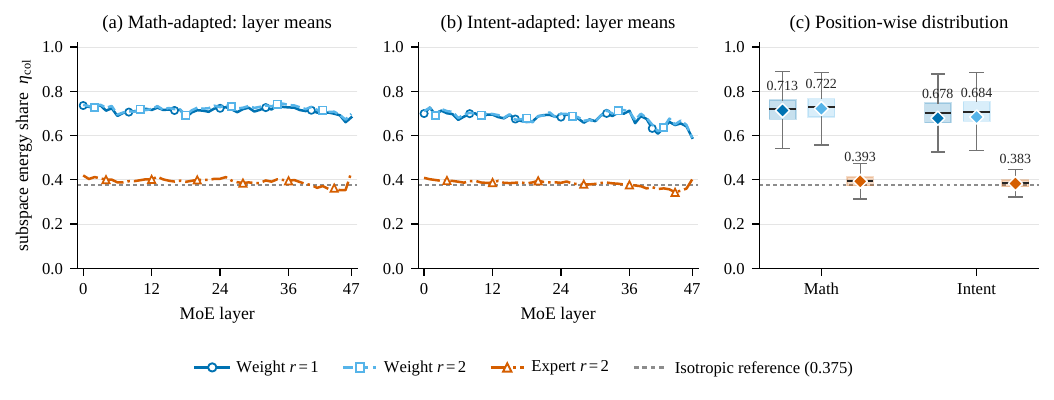}
\caption{Perturbation energy in the pretrained expert output subspace on
Qwen3 synthetic probes.  The left and middle panels show layer means over
128 experts; the right panel shows distributions over 6,144 layer--expert
positions per configuration.  Boxes show medians and interquartile ranges,
whiskers extend to the most extreme observations within 1.5 interquartile
ranges, and diamonds mark means.  The dashed line is the isotropic reference
of 0.375.  Weight-level updates concentrate more energy in pretrained output
subspaces across both checkpoints.}
\label{fig:subspace-energy}
\end{figure}

Figure~\ref{fig:subspace-energy} shows mean $\eta_{\mathrm{col}}$ values of
0.713 and 0.722 for weight-level ranks one and two on Math, and 0.678 and
0.684 on Intent.  Expert-level rank two reaches 0.393 and 0.383,
respectively.  Rank-one weight updates increase the in-subspace energy share
by 32.0 percentage points on Math and 29.5 points on Intent, with a positive
difference in all 48 layer means on both checkpoints.  The rank-one and
rank-two results show that this concentration is already established at
rank one.  Together, $\eta_{\mathrm{col}}$ and $C_{\mathrm{spec}}$ characterize
two complementary aspects of pretrained structure reuse: the energy retained
in the expert's output subspace and its distribution along the pretrained
spectral hierarchy.

\FloatBarrier
\section{Selected Learning Rates and Training Throughput on Qwen3}
\label{app:qwen3-training-throughput}

Table~\ref{tab:qwen3-training-throughput} complements the four design
studies with the selected learning rates and training throughput on
Qwen3-30B-A3B-Base.  Each configuration uses the learning rate selected by
mean downstream performance across three seeds from candidates drawn from
$\{10^{-5},10^{-4},10^{-3}\}$; the same rate is used for all three runs.
Throughput is measured under the
four-GPU training protocol in Appendix~\ref{app:protocol}, with an effective
batch size of 144.  Samples/s measures training examples processed per
second, and steps/s measures optimizer updates per second.

The throughput results support the practical appeal of expert-weight LoRA.
At the selected rates of $10^{-3}$ for Intent and $10^{-4}$ for Math,
weight-level LoRA processes 28.27 and 19.97 samples/s, respectively,
giving $1.31\times$ and $1.45\times$ the throughput of DoRA in these
settings.  It also achieves $1.53\times$ and $1.41\times$ the throughput
of expert-level LoRA on the two tasks.  These comparisons support standard
expert-weight LoRA as an efficient training recipe for both Qwen3 tasks.

\begin{table}[!ht]
\centering
\footnotesize
\setlength{\tabcolsep}{3pt}
\renewcommand{\arraystretch}{1.08}
\begin{tabular}{@{}lcrrcrr@{}}
\toprule
& \multicolumn{3}{c}{Intent} & \multicolumn{3}{c}{Math} \\
\cmidrule(lr){2-4}\cmidrule(l){5-7}
Configuration & LR & Samples/s & Steps/s & LR & Samples/s & Steps/s \\
\midrule
\multicolumn{7}{@{}l}{\textit{Adaptation position}} \\
\addlinespace[2pt]
$A_{64}$ & $10^{-3}$ & $72.034 \pm 3.596$ & $0.504 \pm 0.025$ & $10^{-5}$ & $30.903 \pm 0.298$ & $0.215 \pm 0.002$ \\
$E_{256}$ & $10^{-4}$ & $44.719 \pm 1.206$ & $0.313 \pm 0.008$ & $10^{-5}$ & $29.615 \pm 0.124$ & $0.206 \pm 0.001$ \\
$A_{32}+E_{128}$ & $10^{-4}$ & $38.396 \pm 1.004$ & $0.269 \pm 0.007$ & $10^{-5}$ & $27.960 \pm 0.317$ & $0.194 \pm 0.002$ \\
$A_{64}+R$ & $10^{-4}$ & $70.154 \pm 1.695$ & $0.491 \pm 0.012$ & $10^{-5}$ & $30.835 \pm 0.061$ & $0.214 \pm 0.001$ \\
$E_{256}+R$ & $10^{-4}$ & $43.071 \pm 1.033$ & $0.302 \pm 0.007$ & $10^{-5}$ & $29.129 \pm 0.508$ & $0.202 \pm 0.003$ \\
$A_{32}+E_{128}+R$ & $10^{-4}$ & $38.363 \pm 0.822$ & $0.269 \pm 0.006$ & $10^{-5}$ & $27.360 \pm 0.102$ & $0.190 \pm 0.001$ \\
$A_{64}+E_{256}$ & $10^{-4}$ & $38.008 \pm 0.723$ & $0.266 \pm 0.005$ & $10^{-5}$ & $27.951 \pm 0.361$ & $0.194 \pm 0.002$ \\
$A_{64}+E_{256}+R$ & $10^{-4}$ & $37.875 \pm 0.893$ & $0.265 \pm 0.006$ & $10^{-5}$ & $27.189 \pm 0.108$ & $0.189 \pm 0.001$ \\
\midrule
\multicolumn{7}{@{}l}{\textit{Adaptation granularity}} \\
\addlinespace[2pt]
Module Level ($r=256$) & $10^{-4}$ & $43.664 \pm 1.077$ & $0.306 \pm 0.008$ & $10^{-5}$ & $29.242 \pm 0.683$ & $0.203 \pm 0.005$ \\
Expert Level ($r=2$) & $10^{-3}$ & $18.471 \pm 0.243$ & $0.129 \pm 0.002$ & $10^{-4}$ & $14.123 \pm 0.708$ & $0.098 \pm 0.005$ \\
Weight Level ($r=1$) & $10^{-3}$ & $28.266 \pm 1.027$ & $0.198 \pm 0.007$ & $10^{-4}$ & $19.969 \pm 0.070$ & $0.139 \pm 0.001$ \\
Weight Level ($r=2$) & $10^{-3}$ & $28.986 \pm 0.543$ & $0.203 \pm 0.004$ & $10^{-4}$ & $19.811 \pm 0.138$ & $0.138 \pm 0.001$ \\
\midrule
\multicolumn{7}{@{}l}{\textit{PEFT method}} \\
\addlinespace[2pt]
Weight-level LoRA & $10^{-3}$ & $28.266 \pm 1.027$ & $0.198 \pm 0.007$ & $10^{-4}$ & $19.969 \pm 0.070$ & $0.139 \pm 0.001$ \\
DoRA & $10^{-3}$ & $21.630 \pm 0.995$ & $0.152 \pm 0.007$ & $10^{-4}$ & $13.736 \pm 0.060$ & $0.096 \pm 0.001$ \\
OFT & $10^{-3}$ & $20.519 \pm 0.270$ & $0.144 \pm 0.002$ & $10^{-3}$ & $3.690 \pm 0.011$ & $0.026 \pm 0.000$ \\
FourierFT & $10^{-3}$ & $8.490 \pm 0.019$ & $0.060 \pm 0.001$ & $10^{-3}$ & $7.250 \pm 0.015$ & $0.050 \pm 0.000$ \\
IA3 & $10^{-3}$ & $62.046 \pm 0.887$ & $0.435 \pm 0.006$ & $10^{-3}$ & $28.988 \pm 0.137$ & $0.201 \pm 0.001$ \\
Expert-level LoRA & $10^{-3}$ & $18.471 \pm 0.243$ & $0.129 \pm 0.002$ & $10^{-4}$ & $14.123 \pm 0.708$ & $0.098 \pm 0.005$ \\
MoE-LoRA & $10^{-3}$ & $45.963 \pm 0.405$ & $0.322 \pm 0.003$ & $10^{-5}$ & $30.232 \pm 0.298$ & $0.210 \pm 0.002$ \\
Module-level LoRA & $10^{-4}$ & $43.664 \pm 1.077$ & $0.306 \pm 0.008$ & $10^{-5}$ & $29.242 \pm 0.683$ & $0.203 \pm 0.005$ \\
DAS-LoRA & $10^{-3}$ & $26.508 \pm 0.604$ & $0.186 \pm 0.004$ & $10^{-4}$ & $19.298 \pm 0.122$ & $0.134 \pm 0.001$ \\
Partial FT (last layer) & $10^{-4}$ & $23.357 \pm 1.326$ & $0.164 \pm 0.009$ & $10^{-3}$ & $19.746 \pm 0.480$ & $0.137 \pm 0.004$ \\
\midrule
\multicolumn{7}{@{}l}{\textit{Auxiliary load-balancing loss}} \\
\addlinespace[2pt]
$\lambda=0$ & $10^{-3}$ & $28.266 \pm 1.027$ & $0.198 \pm 0.007$ & $10^{-4}$ & $19.969 \pm 0.070$ & $0.139 \pm 0.001$ \\
$\lambda=10^{-4}$ & $10^{-3}$ & $25.090 \pm 0.752$ & $0.176 \pm 0.005$ & $10^{-4}$ & $20.116 \pm 0.061$ & $0.140 \pm 0.001$ \\
$\lambda=10^{-3}$ & $10^{-3}$ & $24.806 \pm 0.355$ & $0.174 \pm 0.003$ & $10^{-4}$ & $20.072 \pm 0.142$ & $0.139 \pm 0.001$ \\
$\lambda=10^{-2}$ & $10^{-3}$ & $24.216 \pm 0.679$ & $0.170 \pm 0.005$ & $10^{-4}$ & $19.773 \pm 0.695$ & $0.138 \pm 0.005$ \\
$\lambda=10^{-1}$ & $10^{-3}$ & $24.337 \pm 0.288$ & $0.170 \pm 0.002$ & $10^{-3}$ & $19.703 \pm 0.321$ & $0.137 \pm 0.002$ \\

\bottomrule
\end{tabular}
\caption{Selected learning rates and training throughput on
Qwen3-30B-A3B-Base.  Throughput is reported as mean $\pm$ sample standard
deviation across the runs used for each reported configuration.
In the position block, $A_r$ and $E_r$ denote attention and expert-block
adaptation at rank $r$; $R$ adds router adaptation.}
\label{tab:qwen3-training-throughput}
\end{table}

\FloatBarrier

\end{document}